\documentclass{article} 
\usepackage{iclr2026_conference,times}

\newcommand{\eat}[1]{}

\usepackage{balance}
\usepackage{times}
\usepackage{latexsym}
\usepackage{amsfonts}
\usepackage{amsmath}
\usepackage{colortbl}
\usepackage{epsfig}
\usepackage{xspace}
\usepackage{xcolor}
\usepackage{graphicx}
\usepackage{cite}
\usepackage{comment}
\usepackage{booktabs}
\usepackage{multirow}
\usepackage{framed}
\usepackage{subcaption}
\usepackage{wrapfig}
\usepackage{circledsteps}
\usepackage[utf8]{inputenc}
\usepackage{microtype}
\usepackage{inconsolata}
\usepackage[colorlinks=false, linkcolor=black, urlcolor=magenta, citecolor=black]{hyperref}
\usepackage{siunitx}
\usepackage{titlesec}

\usepackage[ruled,linesnumbered]{algorithm2e}
\usepackage{algpseudocode}

\algnewcommand\algorithmicinput{\textbf{Input:}}
\algnewcommand\algorithmicoutput{\textbf{Output:}}
\algnewcommand\Input{\item[\algorithmicinput]}
\algnewcommand\Output{\item[\algorithmicoutput]}

\algdef{SE}[DOWHILE]{Do}{doWhile}{\algorithmicdo}[1]{\algorithmicwhile\ #1}%
\usepackage{lineno}

\definecolor{green}{RGB}{0,128,0}

\definecolor{yellow}{RGB}{255,200,18}

\newcommand{\bi}{\begin{itemize}}
\newcommand{\ei}{\end{itemize}}

\newcommand{\be}{\begin{enumerate}}
\newcommand{\ee}{\end{enumerate}}
\newcommand{\beqn}{\begin{eqnarray*}}
\newcommand{\eeqn}{\end{eqnarray*}}

\newcommand{\ie}{{\em i.e.,}\xspace}
\newcommand{\eg}{{\em e.g.,}\xspace}

\usepackage{tabu}                      
\usepackage{booktabs}                  
\usepackage{lipsum}                    
\usepackage{mwe}                       

\usepackage{listings}
\usepackage{pifont}
\usepackage{tikz}
\usepackage{enumitem}
\usepackage{natbib}
\usepackage{utfsym} 
\usepackage[most]{tcolorbox}

\usetikzlibrary{shapes,snakes}
\usetikzlibrary{calc}

\definecolor{shadecolor}{RGB}{220,220,220}

\tikzstyle{mybox} = [draw=black, fill=black!5, thick,
    rectangle, rounded corners, inner sep=2pt, inner ysep=8pt]
\tikzstyle{fancytitle} =[fill=black, text=white]

\tcbset{
  stagebox/.style 2 args={
    enhanced,
    on line,
    valign=center,
    colback=#1,
    coltext=#2,
    arc=2pt,
    boxrule=0pt,
    boxsep=0pt,
    left=1pt,
    right=1pt,
    top=1.25pt,
    bottom=1pt,
    fontupper=\sffamily\bfseries\footnotesize,
  }
}

\newcommand{\sys}{\textsc{LiteCoST}\xspace}

\definecolor{mygreen}{RGB}{203,230,204}
\definecolor{mydarkgreen}{RGB}{100,230,100}
\definecolor{myred}{RGB}{255,182,193}
\definecolor{mygray}{RGB}{230,230,230}
\definecolor{myblue}{RGB}{138,180,189}
\definecolor{myyellow}{RGB}{255,185,84}
\definecolor{radargreen}{HTML}{38ADA9}

\definecolor{stageAcolor}{HTML}{90b3bd}
\definecolor{stageBcolor}{HTML}{738348}

\usepackage{xcolor}

\definecolor{DeltaUpBg}{HTML}{E6F4EA}
\definecolor{DeltaUpFg}{HTML}{137333}
\definecolor{DeltaDnBg}{HTML}{FDE7E9}
\definecolor{DeltaDnFg}{HTML}{A50E0E}

\newcommand{\SmallBadge}[3]{%
  \textsuperscript{%
    \begingroup
      \setlength{\fboxsep}{0.6pt}
      \scriptsize
      \colorbox{#1}{\textcolor{#2}{\raisebox{0.15ex}{#3}}}%
    \endgroup
  }%
}

\newcommand{\DeltaUp}[1]{\SmallBadge{DeltaUpBg}{DeltaUpFg}{$\uparrow$#1}}
\newcommand{\DeltaDown}[1]{\SmallBadge{DeltaDnBg}{DeltaDnFg}{$\downarrow$#1}}

\usepackage{amsmath,amsfonts,bm}

\def\eqref#1{equation~\ref{#1}}

\def\1{\bm{1}}

\DeclareMathAlphabet{\mathsfit}{\encodingdefault}{\sfdefault}{m}{sl}
\SetMathAlphabet{\mathsfit}{bold}{\encodingdefault}{\sfdefault}{bx}{n}

\title{Long-Document QA with Chain-of-Structured-Thought and Fine-Tuned SLMs}

\author{
Zhuowen Liang$^{1}$,
Xiaotian Lin$^{1}$,
Zhengxuan Zhang$^{1}$,
Yuyu Luo$^{1}$,
Haixun Wang$^{2}$,
Nan Tang$^{1} \thanks{Corresponding author: Nan Tang (E-mail: nantang@hkust-gz.edu.cn)}$
 \vspace{.5em} 
  \\
$^{1}$The Hong Kong University of Science and Technology (Guangzhou),  $^{2}$EvenUp, USA \\
}

\usepackage{hyperref} 
\usepackage{etoc}     

\iclrfinalcopy 
\begin{document}
\maketitle

\begin{abstract}
Large language models (LLMs) are widely applied to data analytics over documents, yet direct reasoning over long, noisy documents remains brittle and error-prone. Hence, we study document question answering (QA) that consolidates dispersed evidence into a structured output (\eg a table, graph, or chunks) to support reliable, verifiable QA. We propose a two‑pillar framework, \textbf{\sys}, to achieve both high accuracy and low latency with small language models (SLMs). 
\textbf{Pillar 1: Chain-of-Structured-Thought (CoST).} We introduce a CoST template, a schema-aware instruction that guides a strong LLM to produce both a step-wise CoST trace and the corresponding structured output. The process induces a minimal structure, normalizes entities/units, aligns records, serializes the output, and verifies/refines it, yielding auditable supervision. 
\textbf{Pillar 2: SLM fine‑tuning.} The compact models are trained on LLM-generated CoST data in two stages: Supervised Fine-Tuning for structural alignment, followed by Group Relative Policy Optimization (GRPO) incorporating triple rewards for answer/format quality and process consistency. By distilling structure-first behavior into SLMs, this approach achieves LLM‑comparable quality on multi-domain long‑document QA using 3B/7B SLMs, while delivering 2–4$\times$ lower latency than GPT‑4o and DeepSeek‑R1 (671B). The code is available at \url{https://github.com/HKUSTDial/LiteCoST}.
\end{abstract}

\section{Introduction}
\label{sec:intro}
\vspace{-0.5em}

\begin{figure}[t!]
\vspace{-2em}
    \centering
    \includegraphics[width=\linewidth]{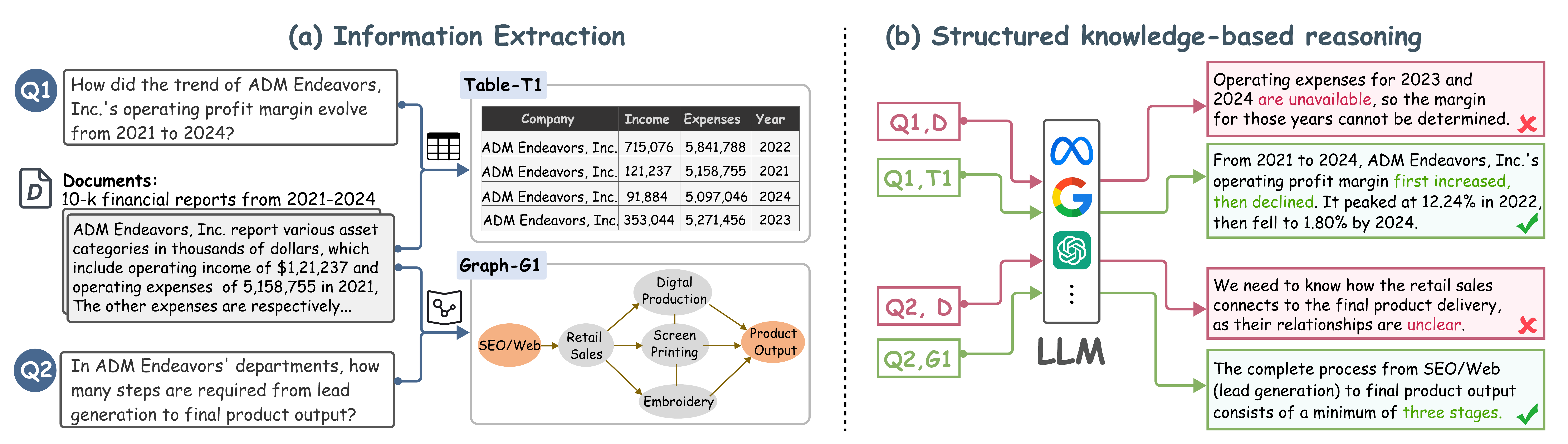}
    \vspace{-2em}
    \caption{
    Structured data makes QA more accurate and reliable. 
    (a) From raw document $D$, we extract a table $T_1$ for query $Q_1$ and a graph $G_1$ for query $Q_2$. 
    (b) LLMs often fail when reasoning directly over unstructured text ($Q_1,D$; $Q_2,D$), but succeed with structured inputs ($Q_1,T_1$; $Q_2,G_1$).
    }
    \label{fig:da_example}
   \vspace{-1.5em}
\end{figure}

Large language models (LLMs) are increasingly used for analytics~\citep{DBLP:conf/cidr/ChenG0FM023,DBLP:journals/corr/abs-2510-23587}, yet direct reasoning over long documents is brittle and opaque, prone to errors in high‑stakes domains such as finance and legal~\citep{chew2023llm,DBLP:journals/corr/abs-2405-12819,edge2024local,DBLP:conf/cidr/0001YF0LH24}. We therefore study long‑document QA where explicit structured data helps.
In this regime, the system constructs a query‑specific structured data—\eg a table, graph, or chunks—from which final answer is directly derivable with explicit explanations. 
As shown in Fig.~\ref{fig:da_example}, extracting structured data for long-document QA~\citep{edge2024local,zhang2025datamosaic,DBLP:conf/iclr/LiC0L0T0H0L25} improves reliability, interpretability, and reuse by exposing evidence and enabling routine verification.

We instantiate a query‑conditioned pipeline: given a natural question $Q$ and documents $D$, the system induces a minimal schema tailored to $Q$, populates it with normalized evidence (\eg units, entities, time), and serializes it into a structured output $S$. The answer $A$ is then computed from $S$. Unlike fixed and pre‑defined schemas, structures are assembled dynamically for each query, thereby excluding open‑ended narrative questions that are not amenable to structured representations.

\begin{figure}[t!]
    \centering
    \includegraphics[width=\linewidth]{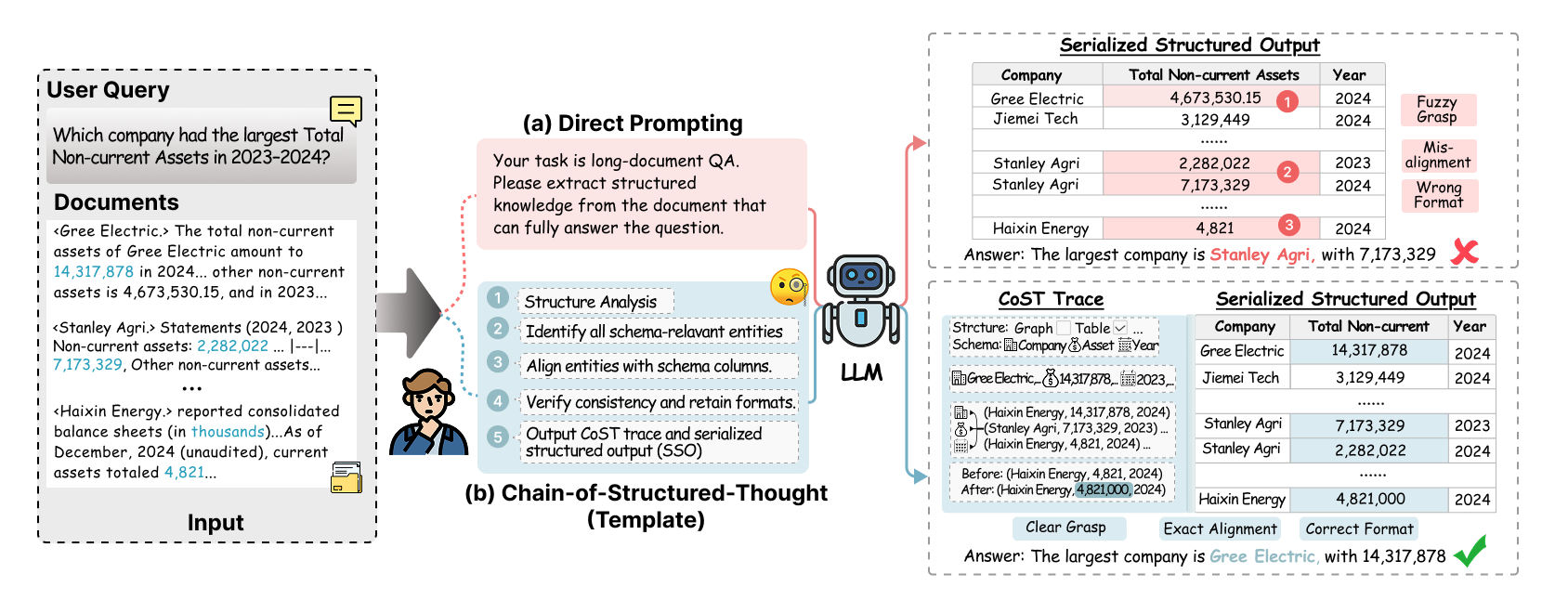}
    \vspace{-2.5em}
    \caption{ 
    (a) Direct prompting LLMs often causes hallucinations and format errors. 
    (b) (Question, Document, CoST Template) $\Rightarrow$ LLM $\Rightarrow$ (CoST Trace, SSO), yielding verifiable and auditable QA.
    }
    \label{fig:da_example2}
   \vspace{-1.5em}
\end{figure}

A natural idea is to directly leverage powerful LLMs (\eg GPT-4 or DeepSeek-R1) to emit the structured artifact. However, \emph{direct prompting is not ideal}: (1) evidence is dispersed across long, multi-document contexts, leading to omissions or hallucinations; (2) values appear in heterogeneous units and formats, requiring normalization; and (3) long-context reasoning must remain consistent across the entire structure. 
As shown in Fig.~\ref{fig:da_example2}(a), direct prompting often yields brittle results—omissions, hallucinations, and format drift~\citep{wei2023zero,DBLP:conf/naacl/WangSLOWZLWG25}. In contrast, Fig.~\ref{fig:da_example2}(b) illustrates our \textbf{CoST template}, which guides the LLM to produce both (i) a schema-aligned \textbf{CoST trace} and (ii) a query-specific \textbf{serialized structured output (SSO)} (\eg table/graph/chunks), ensuring field completeness and format consistency, for robust and interpretable analytics over long-document QA.

%

While CoST prompts with strong LLMs can yield accurate, verifiable SSOs, this effectiveness comes with a substantial cost: repeated large‑model calls increase token/compute budgets, add latency, and limit throughput—undesirable for practical deployments that require low‑latency, high‑throughput service~\citep{DBLP:journals/pvldb/LiLCLT24,xu2024large}. Reliance on hosted LLM APIs can also introduce privacy concerns for sensitive data. A natural response is to adopt small language models (SLMs)\footnote{We use \emph{small language models} (SLMs) to denote compact models (\eg 3B–7B).} 
for on-premises (on-prem) deployment, enabling cost-efficient inference in local or private environments; however, off‑the‑shelf SLMs struggle with the very skills CoST demands, including schema‑aware extraction across long contexts, unit/entity normalization, record alignment, and step‑consistent serialization, making naive LLM$\to$SLM substitution ineffective~\citep{tang2024struc,DBLP:conf/icml/Li0FXC0L25,DBLP:journals/corr/abs-2510-17586}.

To balance \emph{effectiveness} with \emph{efficiency}, we introduce \textbf{\sys}, a two-pillar framework that equips SLMs with strong QA-by-structuring capabilities.
\textbf{Pillar~1} invokes a powerful LLM once as a \emph{structure-first trace generator}: it proposes a concise, query-conditioned schema and produces an auditable CoST trace together with structured data that makes evidence and formats explicit.
\textbf{Pillar~2} \emph{transfers} this ability to an SLM via a lightweight adaptation pipeline: supervised fine-tuning (SFT) to instill structure, format, and step discipline, followed by group-relative policy optimization (GRPO) that jointly rewards answer quality and process consistency.

\textbf{Contributions.} We propose \sys with three notable contributions:
\begin{enumerate}[leftmargin=1.5em,itemsep=-1pt,topsep=2pt]
\item \textbf{CoST for QA-by-Structuring:} a structure-first prompting paradigm that leverages LLMs to elicit step-wise, schema-guided CoST traces and SSOs from long, noisy documents—yielding auditable supervision and machine-checkable outputs.

\item \textbf{SLM adaptation via structured dual signals:} 
a two-phase SFT$\rightarrow$GRPO recipe that introduces a novel dual-level reward, encompassing structured output quality as well as process consistency, to instill CoST-style, schema-aware structured reasoning into compact models.

\item \textbf{Empirical validation:} We have evaluated our approach across multiple domains, including finance, legal, and scientific literature. 
On the \textit{financial} subset of the Loong benchmark~\citep{DBLP:conf/emnlp/WangCCL0WYXZLLY24}, our CoST$\rightarrow$SLM recipe substantially improves small models: LLaMA-3B gains \textbf{+27.6 accuracy points} and \textbf{+0.29 perfect rate (\textit{PR})}, while Qwen-7B gains \textbf{+17.8 accuracy} and \textbf{+0.22 \textit{PR}}, with the 7B model slightly surpassing GPT-4o. Inference is  \textbf{\emph{2–4$\times$ faster}} than GPT-4o/DeepSeek-R1. Extensive experiments further show that our \sys framework delivers significant improvements while showcasing excellent generalization capabilities.
\end{enumerate}

\section{Preliminary and Problem Formulation}
\label{sec:problem}
\vspace{-0.5em}

\textbf{Long-Document QA as \emph{QA-by-Structuring}.}
We study a practically important regime of long-document QA in which the system turns a question $Q$ and a collection of long, noisy, multi-source documents $D$ into a compact, query-specific \emph{serialized structured output} (SSO) with provenance. Structure becomes the interface: the system first induces a minimal schema tailored to $Q$, populates it with normalized, aligned evidence, and then derives the final answer from the structure with explicit support. This framing mitigates noise and dispersion, improves interpretability and reuse, and foregrounds four desiderata: \emph{accuracy} (correct answers), \emph{faithfulness} (evidence-grounded), \emph{auditability} (verifiable traces), and \emph{efficiency} (bounded compute/token cost).

\textbf{Chain-of-Structured-Thought (CoST).}
In our design, the \emph{\textbf{CoST template}} is the \emph{input} to a language model: a schema-aware instruction that specifies step-wise, structure-first requirements.  
When executed, the language model produces two complementary \emph{outputs}:  
(i) a \emph{\textbf{CoST trace}}, \ie the auditable, step-wise reasoning record that documents schema selection, evidence alignment, normalization, and verification; and  
(ii) a \emph{\textbf{serialized structured output}} (SSO), \ie the machine-checkable artifact (table, graph, list, or record set) linked with provenance to the source documents.  
This input–output separation ensures that the LLM’s role is well-defined: given a CoST template, it must emit both a reasoning trace and a structured output, enabling supervision, verification, and reuse.
The whole \textbf{CoST} procedure consists of four key steps: (A1) structure analysis, (A2) trace generation, (A3) quality verification, and (A4) iterative refinement (see Sec.~\ref{ssec: cost} for more details).

\textbf{Two Research Goals.} Our research goals are as follows.

(G1) \textit{Accurate and verifiable QA.} Obtain \emph{high-quality CoST traces and SSOs}—\ie schema-complete, format-consistent, and provenance-grounded outputs—from which we can compute correct answers. 

(G2) \textit{Low latency via SLMs.} Achieve \emph{CoST-style reasoning at SLM speeds}. While strong LLMs are effective CoST generators, their latency/cost hinder deployment. Our objective is to transfer this structure-first behavior to compact models (SLMs) through fine-tuning, so that SLM-generated structures $S_{\text{SLM}}$ are as useful for answering as their LLM counterparts $S_{\text{LLM}}$, at much lower latency:
\begin{equation}
\label{eq:desideratum}
\small
\text{LLM}(Q, S_{\text{SLM}}) \;\approx\; \text{LLM}(Q, S_{\text{LLM}})
\quad \text{and} \quad
\text{Latency}(S_{\text{SLM}}) \;\ll\; \text{Latency}(S_{\text{LLM}}).
\end{equation}
Operationally, we will (i) use LLMs once to generate high-quality CoST traces/SSOs (Pillar~1) and (ii) \emph{fine-tune SLMs} to internalize schema/format/step discipline and process consistency (Pillar~2), enabling accurate, auditable QA at low cost. 

\section{\sys: From LLM CoST Generation to SLM Adaptation}
\label{sec:method}
\vspace{-0.5em}

Next, we present \sys (see Fig.~\ref{fig:framework}), a two-stage framework designed to achieve the dual goals in Sec.~\ref{sec:problem}: (G1) accurate and verifiable QA through high-quality CoST traces and SSOs, and (G2) low-latency execution via compact SLMs. In \textbf{Stage A}, a strong LLM executes the CoST template as input and produces auditable \emph{CoST traces} and machine-checkable \emph{SSOs} as outputs. These outputs serve as supervision signals that capture schema, normalization, alignment, and verification. In \textbf{Stage~B}, we \emph{transfer} this structure-first reasoning behavior into an SLM through a lightweight two-phase recipe: supervised fine-tuning (SFT) for schema/format/step compliance, followed by group-relative policy optimization (GRPO) to jointly reward answer quality and process consistency.

\subsection{Stage A (G1): CoST (Structure-First Reasoning and Trace Generation)}
\label{ssec: cost}
\vspace{-0.5em}

\begin{figure}
\vspace{-2em}
    \centering
    \includegraphics[width=\linewidth]{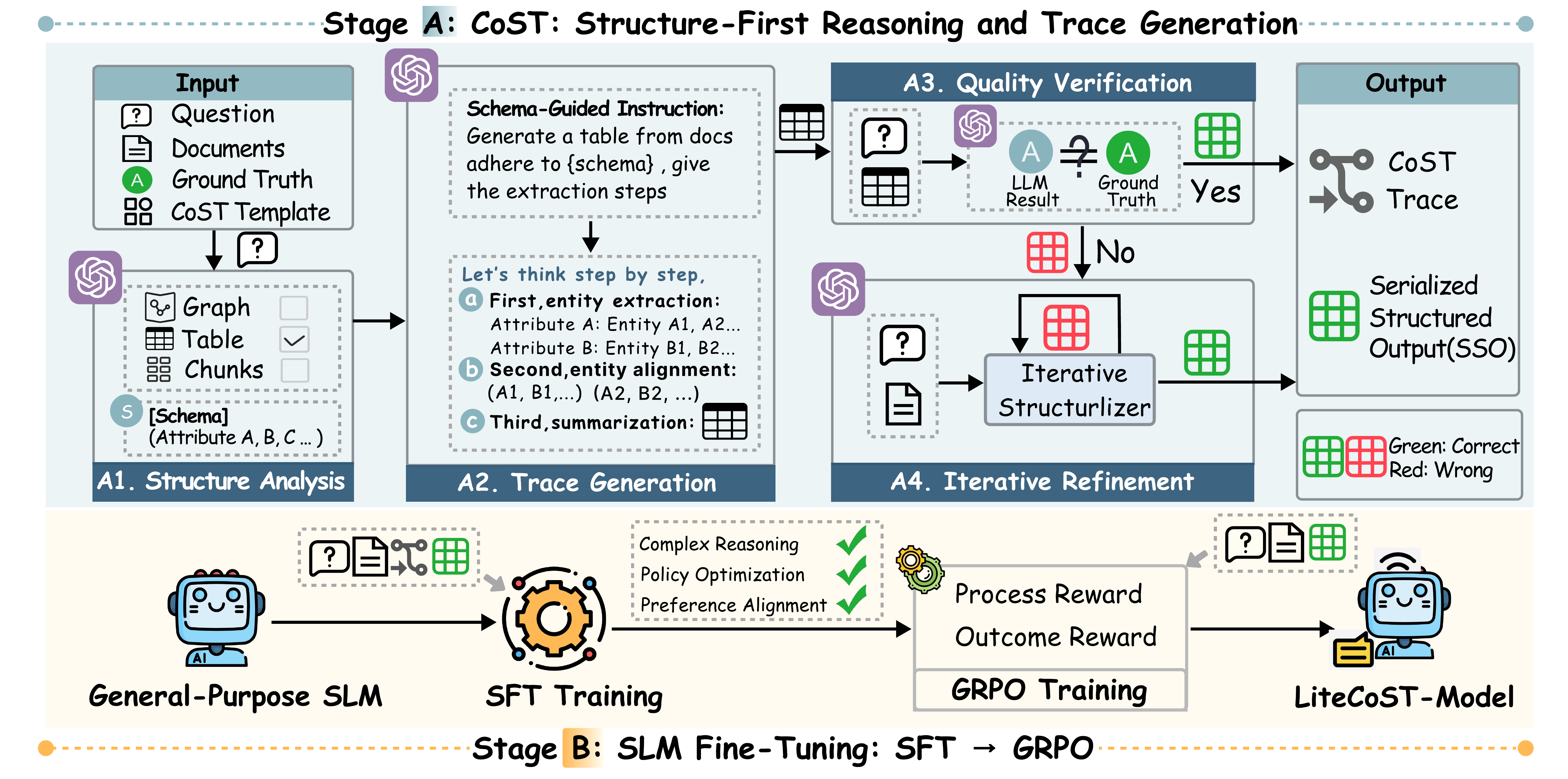}
    \vspace{-2em}
    \caption{Overview of \sys, containing two stages: (1) CoST: Structure-First Reasoning and Trace Generation through structure analysis, trace generation, quality verification, and iterative refinement; and (2) SLM Fine-Tuning: SFT → GRPO process, including SFT for structure/format/steps, followed by GRPO with dual signals for answer/format quality and process consistency.}
    \label{fig:framework}
    \vspace{-1em}
\end{figure}

As illustrated in Fig.~\ref{fig:framework}, we operationalize CoST as a \emph{structure-first, input$\to$output} procedure: given a question, document, the ground truth answer, and CoST template, a strong LLM generates two outputs—an auditable \emph{CoST trace} and a serialized structured output \emph{SSO}.

\textbf{(A1) Structure Analysis.}
%
The first step is to dynamically select the most suitable data structure and instantiate an accurate schema to support answering a given question. Specifically, \sys incorporates a question-oriented structure selection mechanism that, for example, chooses tables for statistical comparison or graphs for relational reasoning, without exhaustively processing the entire corpus.
Once the structure type is chosen, we invoke a dynamic \textbf{schema construction} procedure in which the LLM parses the question and enumerates task-specific attributes/entities (\eg \texttt{Company}, \texttt{Asset}, \texttt{Year}), ensuring precise alignment with the question semantics.

\textbf{(A2) CoST Trace Generation.}
Following structure analysis, we adopt an instruction-based chain-of-thought paradigm that performs step-by-step reasoning to progressively generate the trace to guide schema-aligned extraction.
The task is specified through three key components: 1) \textbf{task description}, a template specifying step-wise requirements; 2) \textbf{input text}, the source documents; and 3) the dynamically generated \textbf{schema}.
Guided by schema-informed instructions, a strong LLM extracts, aligns, and serializes into a deterministic structured format, emitting both the reasoning trace and the final structured output.
The template of trace generation is provided in Appendix~\ref{appendix:trace_generation}.

\textbf{(A3) Quality Verification.} The module aims to assess the quality of the generated structured data by evaluating its ability to answer the original question. 
Since ground-truth structured data is unavailable, we adopt an LLM-as-Judge approach~\citep{zheng2023judging}, where a strong LLM evaluator (e.g., GPT-4o) assesses the extracted responses. Inference outputs that exactly match the reference answers are deemed correct and retained for subsequent training.
Further details are provided in Appendix~\ref{appendix: data_verification}.

\textbf{(A4) Iterative Refinement.}
At its core, the module employs an \textbf{Iterative Structuralizer} that refines low-quality samples by regenerating structured knowledge for GRPO training. Rather than discarding flawed but challenging cases, it reuses them recursively with the question and context, reframing the task as supplemental extraction and providing richer supervision than vanilla fine-tuning. The iterative update rule, sufficiency evaluator, and stopping criteria are detailed in Appendix~\ref{appendix: data_refinement}.





\textbf{Final Output.}
After the CoST pipeline, the final output is $(c^*, S^*)$, where $c^*$ is the CoST trace (generated in A2), and $S^*$ the structured output refined through quality verification (A3) and iterative refinement (A4). This pair provides high-quality supervision for training and downstream reasoning.

\vspace{-0.5em}
\subsection{Stage B (G2): SLM Fine-Tuning (SFT $\rightarrow$ GRPO)}
\vspace{-0.5em}

Given the supervised training data, \sys first warms up the model with Supervised Fine-Tuning (SFT). We then apply reinforcement learning with Group Relative Policy Optimization (GRPO), introducing a dual-level reward that jointly optimizes (1) outcome reward, which evaluates the format
compliance and answer correctness, and (2) process reward, which scores step-wise reasoning against ground-truth evidence to enforce a reliable extraction path.

\textbf{Training Data Template.} 
Each training sample is defined as 
$z = (i, d, c^*, y^*)$, 
where $i$ is the question, $d$ the document input, $c^*$ the CoST reasoning trace (enclosed by \texttt{<reasoning>...</reasoning>}), and $y^*$ the structured output (enclosed by \texttt{<answer>...</answer>}). 
In \emph{SFT}, the model learns to generate $(c^*, T^*)$ from $(i, d)$, 
while \emph{GRPO} also conditions on $(i, d)$ and optimizes with dual-level rewards (process, outcome) against the verified targets.

\textbf{Supervised Fine-tuning (SFT).}
We initially performed Supervised Fine-Tuning (SFT) on a general-purpose base model, specifically enhancing its capability for CoT-driven information extraction. This process enables the model to acquire fundamental extraction capability (\eg handling structure, format, and step-wise reasoning) in a specific domain, thus substantially mitigating the errors observed when deploying the base model on complex extraction tasks.

\begin{figure}
\vspace{-2em}
    \centering
    \includegraphics[width=\linewidth]{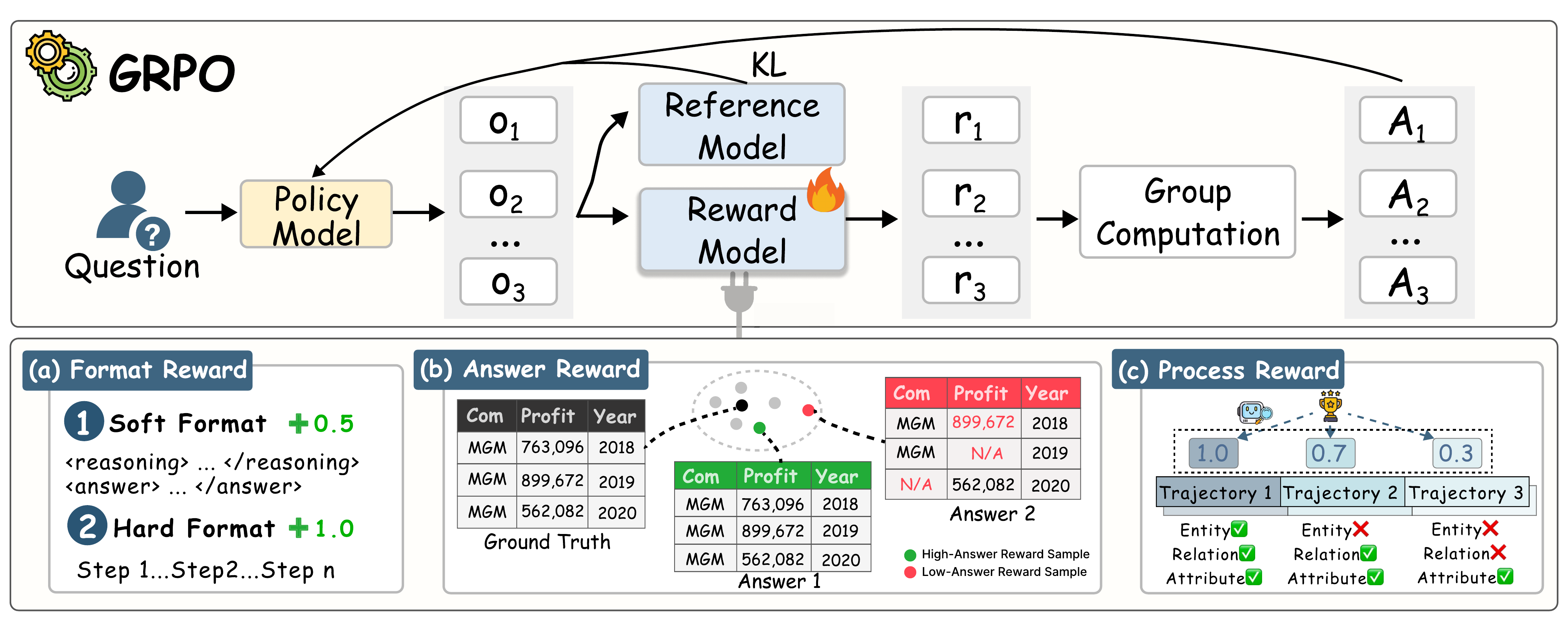}
    \vspace{-2em}
    \caption{The GRPO training pipeline based on dual-level reward.}
    \label{fig:grpo}
    \vspace{-0.5em}
\end{figure}

\textbf{Group Relative Policy Optimization (GRPO).}
We then employ GRPO via a three-level reward mechanism, as illustrated in Fig.~\ref{fig:grpo}.

\underline{\textit{Formulation.}} For each question \( q \), GRPO samples a group of outputs \( \{o_1, o_2, \dots, o_G\} \) from the old policy $\pi_\theta$. Each output receives a reward \( r_i \), yielding a set of \( G \) rewards \( \mathbf{r} = \{r_1, r_2, \dots, r_G\} \). From these rewards, we compute the group-relative advantage
$A_{i} = \frac{r_{i} - \mathrm{mean}(r_{1} \ldots r_{G})}{\mathrm{std}(r_{1} \ldots r_{G})}$
and then optimizes the policy model by maximizing the following objective:
\begin{equation}
\begin{aligned}
&\mathcal{J}_{\mathrm{GRPO}}(\theta) = \mathbb{E}
\left[ 
\mathbf{v} \sim P(\mathbf{V}), \{o_i\}_{i=1}^G \sim \pi_{\theta_{\mathrm{old}}}(\mathcal{O}|\mathbf{v})
\right] \\[-0.6em] 
&\quad \frac{1}{G} \sum_{i=1}^{G} \left( \min \left( r_i^{\mathrm{ratio}} A_i, \mathrm{clip} \left( r_i^{\mathrm{ratio}}, 1 - \epsilon, 1 + \epsilon \right) A_i \right) - \beta D_{\mathrm{KL}}(\pi_\theta \| \pi_{\mathrm{ref}}) \right),
\end{aligned}   
\end{equation}

\vspace{-1em}
where $\beta$ and $\epsilon$ are hyper-parameters, $r^{\text{ratio}}_i$ is the importance sampling ratio comparing the likelihood of output $o_i$ under the new and old policies, and $A_i$ is the group-relative advantage. The clipping operator would stabilize updates within a trust region, and the minimum operation ensures conservative yet effective policy updates~\citep{shao2024deepseekmath}. Further details are provided in Appendix~\ref{appendix:grpo}.


\underline{\textit{Format Compliance.}}
Fig.~\ref{fig:grpo} (a) illustrates the hierarchical design of the format reward: a soft reward (0.5) is given if the output contains a single reasoning sequence in \texttt{<reasoning>} and a final answer in \texttt{<answer>} without extraneous content; a hard reward (1.0) is assigned if the reasoning is further structured with explicit step labels (\eg Step 1, Step 2); otherwise, the score is 0.
%


\underline{\textit{Answer Correctness.}} 
As illustrated in Fig.~\ref{fig:grpo}(b),  we address the limitations of rule-based evaluation by adopting a hybrid metric that combines structural alignment and semantic similarity:
\begin{equation}
\label{answer_award}
\small
f_{\text{score}} = \alpha \cdot \mathcal{S}_{\text{struct}} + (1 - \alpha) \cdot \mathcal{S}_{\text{sem}}
\end{equation}
For $\mathcal{S}_{\text{struct}}$, we use rule-based checks (\eg row-column alignment in tables) to verify structural correctness. For $\mathcal{S}_{\text{sem}}$, we adopt GPT-4o-mini as an automatic evaluator, comparing the content within \texttt{<answer>...</answer>} tags against the reference; outputs with higher semantic similarity receive higher rewards. The raw score $f_{\text{score}}$ is scaled from $[0,100]$ to $[0,1]$, with NULL rewards assigned to empty outputs. The detailed LLM-based evaluation prompts are provided in Appendix~\ref{appendix:answer_completeness}.


\underline{\textit{Process Reward.}} 
Outcome rewards alone are sparse and insufficient for fine-grained guidance. We therefore introduce a consistency-based process reward to supervise reasoning at the step level.
Consistency is evaluated from both the entity-level and the tuple-level, enabling the model to capture fine-grained errors such as partially incorrect entities or mismatched relations. For each step $i$, LLM is prompted with an instruction $I_{\text{consistency}}$ to judge whether the predicted step result $s_i$ is consistent with the corresponding ground truth $s_i^*$. If the consistency holds, the step is assigned a score of $1$; otherwise, it receives $0$. The overall process reward is formally defined as:
%
%
\begin{equation}
\small
R_{\text{process}}(s_i) \;=\; \frac{1}{N} \sum_{i=1}^{N} \mathbf{1}\!\left[ \text{Cons}(s_i, s_i^* \mid I_{\text{consistency}}) \right],
\end{equation}
\vspace{-1em}


where $N$ denotes the total number of reasoning steps, $\text{Cons}(\cdot)$ is the LLM-based consistency function, and $\mathbf{1}[\cdot]$ is the indicator function that returns $1$ if the consistency check is satisfied and $0$ otherwise. This formulation provides a dense and fine-grained training signal, guiding the model towards faithful step-by-step extraction while complementing the sparse outcome reward, as shown in Fig.~\ref{fig:grpo} (c).

\underline{\textit{Overall Reward.}} 
%
The overall reward is defined as the sum of the format compliance, answer correctness, and process rewards. To prevent training dynamics from being dominated by other reward signals, we introduce a scaling factor that modulates the process reward along each trajectory,  $\tilde{R}_{\text{process}}(s_i) = R_{\text{process}}(s_i) \cdot \gamma(T_i)$. Here, $\gamma(T_i)$ is a trajectory-level coefficient: positive for correct answers to reinforce reasoning, negative for incorrect or overthought trajectories to discourage such behaviors, and $1$ for format errors to isolate penalties to specific steps.

\vspace{-0.3em}
\section{Experiments on Long-Document QA with CoST and SLMs}
\label{sec:experiment}
\vspace{-0.5em}


In this section, we evaluate the performance of our proposed \sys framework on the Loong benchmark~\citep{DBLP:conf/emnlp/WangCCL0WYXZLLY24}, which effectively captures the challenges of generating \textit{serialized structured output (SSO)} across varying context lengths. Rather than directly answering questions, we focus on the ability of \sys to produce reliable SSO that supports long-document QA. We further assess its efficiency and conduct ablation studies to analyze contributing factors. Specifically, we aim to address the following research questions:

\vspace{-0.5em}
\begin{description}[noitemsep, topsep=0pt, leftmargin=1.5em]
    \item[\textbf{(1) Benefits of Structured Data:}] How do structured outputs enhance long-document QA?
    \item[\textbf{(2) Effectiveness:}] How effective is \sys in generating high-quality SSO for long-document QA, compared with current LLMs and state-of-the-art methods?
    \item[\textbf{(3) Efficiency:}] How efficient is \sys relative to LLMs in terms of SSO generation speed?
    \item[\textbf{(4) Ablation Study:}] What factors contribute to performance gains in structured output generation?
    \item[\textbf{(5) Generalization:}] How well does the  framework generalize to other datasets and domains?
\end{description}

%


\subsection{Experimental Setup}
\label{ssec:exp_setup}

\textbf{Training Dataset.}
To support two-phase training, we construct two domain-specific datasets via \sys from four large-scale multi-task resources: \textsc{FinQA}~\citep{chen2021finqa}, \textsc{TAT-QA}~\citep{zhu2021tat}, \textsc{Squad}~\citep{rajpurkar2016squad}, and \textsc{LegalBench}~\citep{pipitone2024legalbench}. 
The datasets target the \textit{finance}, \textit{legal} and \textit{general} knowledge, capturing diverse reasoning patterns (e.g., aggregation, comparison, multi-hop inference) in realistic settings drawn from diverse documents.  

\textbf{Evaluation Dataset.}
%
We adopt the Loong benchmark~\citep{DBLP:conf/emnlp/WangCCL0WYXZLLY24}, a real-world multi-document QA dataset with 1,600 test samples spanning three domains (Finance, Legal, Paper), four task categories (Spotlight Locating, Comparison, Clustering, Chain of Reasoning), and four document length settings where longer contexts disperse relevant information. Our analysis focuses primarily on an in-depth analysis of the finance domain, with legal results provided in Appendix~\ref{appendix:legal_qa}.

\textbf{Evaluation Details.} 
Defining ground truth for the structured output from long-context documents, poses significant challenges. To address this, we adopt a 2-hop evaluation paradigm by leveraging downstream QA tasks~\citep{DBLP:conf/acl/JainMP24}. 
 %
Specifically, we employ the GPT-4o as an automatic judge, scoring model responses from 0 to 100 based on accuracy, hallucination, and completeness. It also introduces the Perfect Rate, which measures the proportion of responses that achieve a perfect score. 

\textbf{Baselines.} 
To comprehensively evaluate the generation capability of \sys, we compare the performance gains achieved through both LLMs and SLMs. Specifically, we consider two categories of baselines: the first targets reasoning, where LLMs are prompted to generate structured output (\eg Zero-shot and Chain-of-Thought (CoT)~\citep{DBLP:conf/nips/Wei0SBIXCLZ22}). 
The second focuses on improvements with SLMs, where we compare against several state-of-the-art models, including Llama3.2-3B-Instruct, Qwen2-7B-Instruct, Llama-3.1-8B-Instruct, Qwen2.5-14B-Instruct, GPT4o-mini, GPT-4o, and DeepSeek-R1. We further include two categories of baselines: (1) Fine-tuned IE models,  such as ODIE~\citep{DBLP:conf/emnlp/Jiao0LZOJ023}, IEpile~\citep{gui-etal-2024-iepile}, and Struc-bench~\citep{tang2024struc}; and (2) Modular extraction frameworks that leverage component modules to extract structured knowledge, including StructRAG~\citep{DBLP:conf/iclr/LiC0L0T0H0L25}.
For a fair comparison, we evaluate the
baseline methods using the same backbones (\ie
LLaMA-3.2-3B-Instruct, Qwen2-7B-Instruct) as those used for our \sys.
These baselines act as structured output generator, with GPT-4o employed as the reasoning model to produce responses. More details are provided in Appendix~\ref{appendix:settings}.
%

\textbf{Implementation Details.} 
During the training phase, \sys employs a two-phase training pipeline comprising LoRA fine-tuning followed by GRPO optimization using trl and verl.
The model is trained in two stages: first, fine-tuning for 3 epochs with a learning rate of 2e-4 and batch size of 16 using a LoRA adapter (rank 16, lora alpha 32); followed by reinforcement learning with GRPO using a learning rate of 1e-5, batch size of 16, and 5 sampled generations per query. In Equation~\ref{answer_award}, the weighting parameter $\alpha$ is set to 0.3; the training cost is about $\$20$, and the maximum generation length is extended to 2,048 tokens to support CoT-style reasoning.

\begin{table}[t!]
\vspace{-1em}
\centering
\caption{Comparison of different models generating structured outputs for long-document QA on the \textit{Finance} Subset of Loong. \colorbox{mygreen}{Green} highlights the remarkable improvements over the base model.}
\vspace{-1em}

\label{tab:litesea_results}
\setlength{\tabcolsep}{3pt} 
\small 
\resizebox{\textwidth}{!}{ 
\begin{tabular}{@{} l c *{10}{c} @{}}
\toprule
\multirow{2}{*}{\makebox[1.8cm][l]{\textbf{ Model}}} & \textbf{Model} & \multicolumn{2}{c}{\textbf{\textit{Spotlight Locating}}} & \multicolumn{2}{c}{\textbf{\textit{Comparison}}} & \multicolumn{2}{c}{\textbf{\textit{Clustering}}} & \multicolumn{2}{c}{\textbf{\textit{Chain of Reasoning}}} & \multicolumn{2}{c}{\textbf{\textit{Overall}}} \\
\cmidrule(lr){3-4} \cmidrule(lr){5-6} \cmidrule(lr){7-8} \cmidrule(lr){9-10} \cmidrule(lr){11-12}
 & \textbf{Size} & \textbf{\textit{AS}} & \textbf{\textit{PR}} & \textbf{\textit{AS}} & \textbf{\textit{PR}} & \textbf{\textit{AS}} & \textbf{\textit{PR}} & \textbf{\textit{AS}} & \textbf{\textit{PR}} & \textbf{\textit{AS}} & \textbf{\textit{PR}} \\
\midrule
\multicolumn{12}{>{\columncolor[gray]{.88}}c}{\textit{Close-Sourced Models \& Large Language Models}}  \\
LLaMA-3.1-8B-Instruct & 8B & 55.03 & 0.20 & 51.60 & 0.15 & 51.50 & 0.04 & 44.75 & 0.02 & 51.32 & 0.10 \\
GPT-4o-mini & 8B & 84.42 & 0.70 & 80.40 & 0.67 & 77.38 & 0.40 & 65.35 & 0.18 & 78.08 & 0.51 \\
Qwen2.5-14B-Instruct & 14B & 83.74 & 0.57 & 82.12 & 0.56 & 69.96 & 0.24 & 66.41 & 0.10 & 75.60 & 0.38 \\
GPT-4o~\citep{abacha2025medec} & 200B & 84.10 & 0.73 & 80.53 & 0.60 & 81.50 & 0.50 & 64.30 & 0.25 & 79.32 & 0.54 \\
Deepseek-R1~\citep{guo2025deepseek}  & 671B & 84.27 & 0.62 & 78.97 & 0.55 & 75.42 & 0.34 & 74.40 & 0.35 & 78.18 & 0.46 \\
\midrule
LLaMA-3.2-3B-Instruct (Base)  & 3B & 49.90 & 0.16 & 52.10 & 0.14 & 47.89 & 0.07 & 46.85 & 0.06 & 49.37 & 0.11 \\
LLaMA-3.2-3B-Instruct (\textit{Ours}) & 3B
& \textbf{81.27}
& \textbf{0.53} 
& \textbf{78.08} 
& \textbf{0.49} 
& \textbf{78.34} 
& \textbf{0.36} 
& \textbf{64.75} 
& \textbf{0.16} 
& \textbf{76.95}
& \textbf{0.40} \\
& 
& \DeltaUp{31.37}
& \DeltaUp{0.37}
& \DeltaUp{25.98}
& \DeltaUp{0.35}
& \DeltaUp{30.45}
& \DeltaUp{0.29}
& \DeltaUp{17.90}
& \DeltaUp{0.10}
& \DeltaUp{27.58}
& \DeltaUp{0.29} \\[-5pt]

\midrule
Qwen2-7B-Instruct (Base) & 7B & 63.10 & 0.36 & 67.85 & 0.37 & 60.83 & 0.18 & 52.25 & 0.09 & 62.10 & 0.26 \\
Qwen2-7B-Instruct (\textit{Ours}) & 7B 
& \textbf{83.97}
& \textbf{0.62}
& \textbf{81.55}
& \textbf{0.59}
& \textbf{81.00}
& \textbf{0.43}
& \textbf{67.98}
& \textbf{0.18}
& \textbf{79.93}
& \textbf{0.48} \\
& 
& \DeltaUp{20.87}
& \DeltaUp{0.26}
& \DeltaUp{13.70}
& \DeltaUp{0.22}
& \DeltaUp{20.17}
& \DeltaUp{0.25}
& \DeltaUp{15.73}
& \DeltaUp{0.09}
& \DeltaUp{17.83}
& \DeltaUp{0.22} \\[-5pt]
\bottomrule
\end{tabular}%
}
\vspace{-1em}
\end{table}

\begin{figure}[t]
    \centering
    \begin{minipage}[t]{0.45\linewidth}
        \centering
        \includegraphics[width=\linewidth]{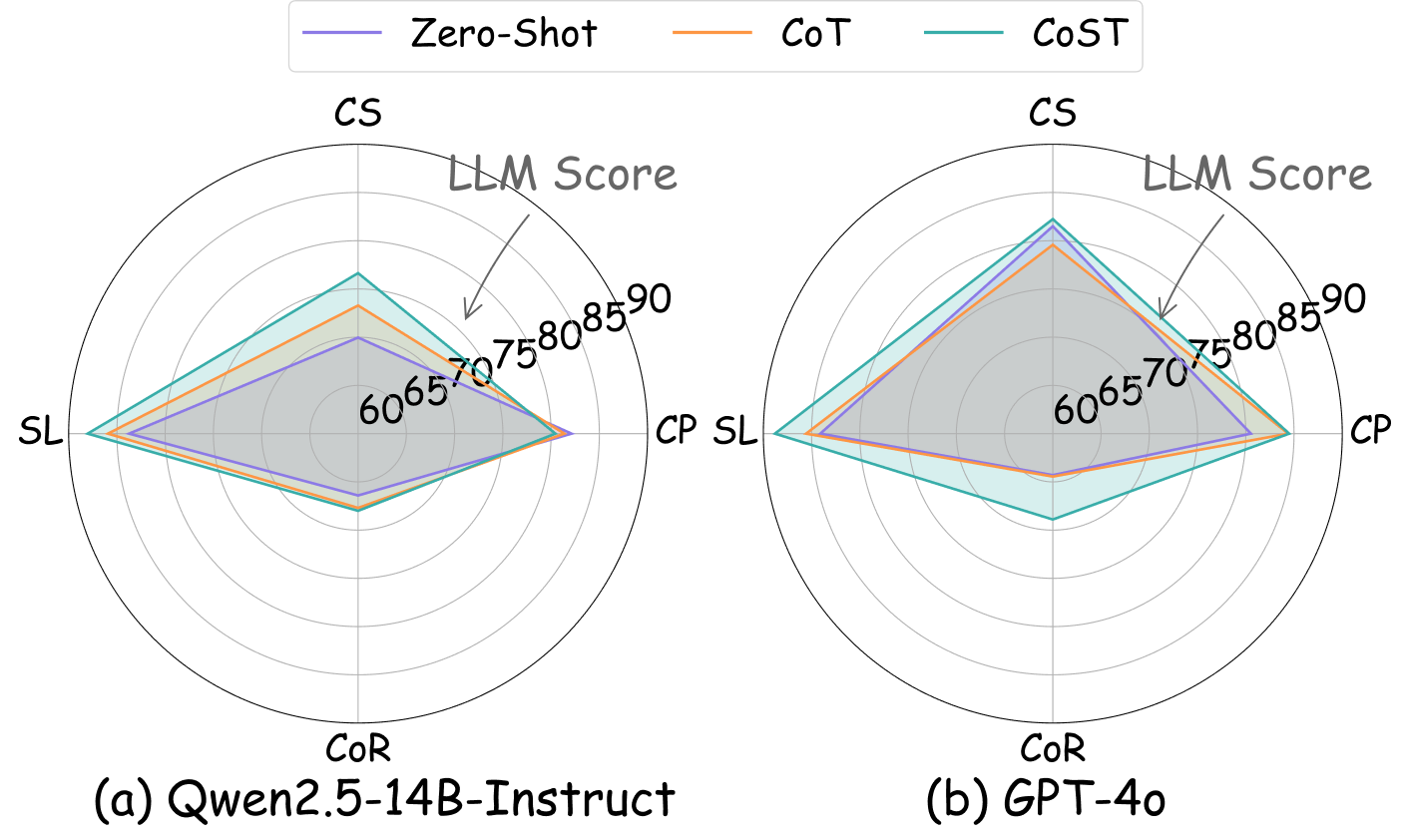}
        \vspace{-1em}
        \captionof{figure}{Radar plot of detailed scores for different prompting methods on 4 subtasks on the \textit{Finance} subset of Loong.}
        \label{fig:llm_enhancement}
    \end{minipage}
    \hfill
    \begin{minipage}[t]{0.54\linewidth}
        \centering
        \includegraphics[width=\linewidth]{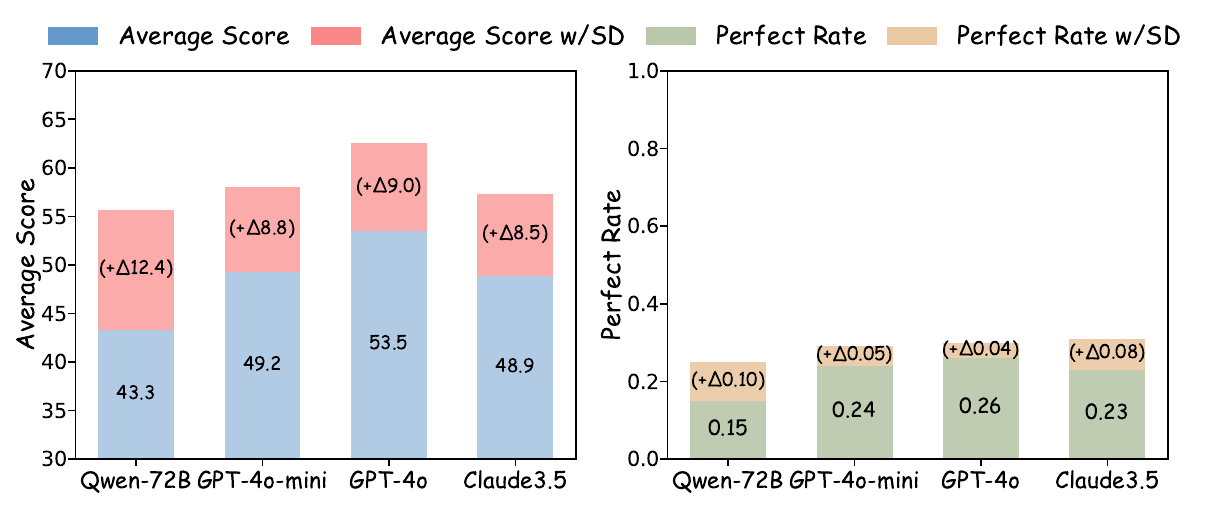}
        \vspace{-1em}
        \captionof{figure}{Quality assessment of CoST-generated structured data via reasoning performance on Loong across popular LLMs (\textit{SD} denotes structured data).}
        \label{fig:ablation_sd}
    \end{minipage}
    \vspace{-1em}
\end{figure}

\vspace{-0.3em}
\subsection{The Benefits of Structured Data}
This experiment evaluates how structured data improves model performance on knowledge-intensive reasoning, emphasizing the need for accurate serialized structured output (SSO). The data is curated by \sys using GPT-4o as the base model; structure distributions are detailed in Appendix~\ref{appendix: Loong_structure}.

\textbf{High-quality SSO from CoST improves LLM Reasoning.}
%
As shown in Fig.~\ref{fig:ablation_sd}, all models achieve stronger reasoning when leveraging the structured data rather than raw long documents. With \sys, overall scores rise by 12.41, 8.77, 9.04, and 8.47 points, and perfect rates improve by +0.10, +0.05, +0.04, and +0.08 for Qwen2-72B-Instruct, GPT-4o-mini, GPT-4o, and Claude-3.5-Sonnet, respectively. These consistent gains highlight the value of high-quality structured knowledge in improving both accuracy and reliability. Full results are provided in Appendix~\ref{appendix: Loong_structure}.

%

\vspace{-0.3em}
\subsection{Effectiveness: How good is \sys for SSO Generation?}
In this section, we evaluate the effectiveness of \sys across both LLMs and compact SLMs on the \textit{Finance} subset of Loong, compared with state-of-the-art models and baseline methods.
The in-depth analysis demonstrate that \sys consistently outperforms other comparable strategies, achieving substantial gains in correctness, and Sec.~\ref{ssec:generalization} further confirms its broader generalization.

\textbf{Efficacy of Chain-of-Structured-Thought.}
Fig.~\ref{fig:llm_enhancement} presents a radar chart comparing the performance of different prompting methods across four task categories, evaluated on two backbone LLMs. 
The results show that step-wise reasoning substantially improves SSO generation and yields high-quality supervision signals, with CoT consistently outperforming Zero-Shot, especially on Qwen2.5-14B-Instruct.
Beyond this, our structured prompting paradigm (CoST) yields the strongest improvements, consistently reaching the outer boundary (\textcolor{radargreen}{green line}) and achieving top performance across nearly all tasks, notably in Chain of Reasoning (GPT-4o), Clustering (Qwen2.5-14B-Ins), and Spotlight Locating, underscoring its consistent effectiveness.
See full numerical results in the Appendix~\ref{appendix:full_radar_results}.

\textbf{Our fine-tuned SLMs $\gg$ other SLMs.}
Table~\ref{tab:litesea_results} demonstrates that our model consistently outperforms all evaluated small language models (defined as open-source models with fewer than 8 billion parameters). 
Both fine-tuned variants, \textbf{LLaMA-LiteCoST} and \textbf{Qwen-LiteCoST}, achieve substantial improvements over their base models, with gains of (+27.58, +0.29) and (+17.83, +0.22), and consistent enhancements across all sub-tasks. Compared with other small scales, the LLaMA-LiteCoST significantly outperforms the 7B, and 8B models by (+14.85, +0.14) and (+25.63, +0.30) in overall score and perfect rate, respectively, while Qwen-LiteCoST delivers even larger improvements, underscoring the effectiveness of our approach.

\textbf{Our fine-tuned SLMs $\approx$ LLMs.}
On one hand, both variants surpass Qwen-14B-Instruct despite having far fewer parameters, with improvements of (+1.35, +0.02) on the LLaMA backbone and (+4.33, +0.10) on the Qwen backbone. 
%
On the other hand, \textbf{Qwen-LiteCoST} achieves the best overall performance among all evaluated models, surpassing all small baselines and even exceeding GPT-4o-mini by 1.85, Deepseek-R1 by 1.75, and GPT-4o by 0.61. It achieves top-2 performance on 5 out of 8 evaluation points across the four subtasks, highlighting the effectiveness of our training strategy in narrowing the gap between lightweight and large-scale models for structured output generation.
Full results across different document sizes are provided in Appendix~\ref{appendix:full_result}.

\textbf{RL-enhanced gains for SLMs.} 
Compared with other state-of-the-art methods, including fine-tuned models and modular extraction frameworks, \sys achieves superior performance across all tasks, as shown in Table~\ref{tab:litesea_baselines}. In particular, \sys substantially outperforms the strong baseline StructRAG, with gains of (+30.91, +0.39) on LLaMA and (+30.47, +0.46) on Qwen, 
Moreover, relative to the previous best fine-tuned methods, it sets a new state of the art, achieving improvements of (+15.05, +0.18) over IEPile on LLaMA and (+6.41, +0.05) over Strucbench on Qwen.
These results indicate that our RL-enhanced framework provides a principled advancement over conventional fine-tuning for information extraction, further underscoring its effectiveness and robustness.

\begin{table}[t!]
\vspace{-1em}
\centering
\caption{Performance of the \textit{Finance} subset of Loong  compared with other state-of-the-art methods }
\vspace{-1em}
\label{tab:litesea_baselines}
\setlength{\tabcolsep}{3pt}
\small
\resizebox{\textwidth}{!}{%
\begin{tabular}{@{} l l *{10}{c} @{}}
\toprule
\multirow{2}{*}{\makebox[2.5cm][l]{\textbf{Backbone}}} 
& \multirow{2}{*}{\makebox[2.5cm][l]{\textbf{Method}}} 
& \multicolumn{2}{c}{\textbf{\textit{Spotlight Locating}}} 
& \multicolumn{2}{c}{\textbf{\textit{Comparison}}} 
& \multicolumn{2}{c}{\textbf{\textit{Clustering}}} 
& \multicolumn{2}{c}{\textbf{\textit{Chain of Reasoning}}} 
& \multicolumn{2}{c}{\textbf{\textit{Overall}}} \\
\cmidrule(lr){3-4} \cmidrule(lr){5-6} \cmidrule(lr){7-8} \cmidrule(lr){9-10} \cmidrule(lr){11-12}
& & \textbf{\textit{AS}} & \textbf{\textit{PR}} & \textbf{\textit{AS}} & \textbf{\textit{PR}} & \textbf{\textit{AS}} & \textbf{\textit{PR}} & \textbf{\textit{AS}} & \textbf{\textit{PR}} & \textbf{\textit{AS}} & \textbf{\textit{PR}} \\
\midrule

\multirow{5}{*}{LLaMA-3.2-3B-Ins} 
& ODIE~\citep{DBLP:conf/emnlp/Jiao0LZOJ023}     & 68.89 & 0.39 & 61.30 & 0.30 & 61.11 & 0.15 & 49.75 & 0.07 & 61.21 & 0.23 \\
& IEpile~\citep{gui-etal-2024-iepile}           & 62.90 & 0.37 & 65.10 & 0.30 & 63.12 & 0.14 & 50.95 & 0.02 & 61.90 & 0.22 \\
& Struc-bench~\citep{tang2024struc}      & 55.13 & 0.15 & 51.05 & 0.15 & 48.16 & 0.06 & 44.10 & 0.08 & 49.90 & 0.11 \\
& StructRAG~\citep{DBLP:conf/iclr/LiC0L0T0H0L25}    & 39.50 & 0.01 & 39.70 & 0.02 & 31.08 & 0.00 & 35.95 & 0.00 & 36.04 & 0.01 \\
& \sys (Ours) & \textbf{81.27} & \textbf{0.53} & \textbf{78.08} & \textbf{0.49} & \textbf{78.34} & \textbf{0.36} & \textbf{64.75} & \textbf{0.16} & \textbf{76.95} & \textbf{0.40} \\
\midrule

\multirow{5}{*}{Qwen2-7B-Ins} 
& ODIE~\citep{DBLP:conf/emnlp/Jiao0LZOJ023}   & 82.13 & 0.59 & 73.85 & 0.48 & 73.54 & 0.32 & 55.30 & 0.12 & 72.86 & 0.40 \\
& IEpile~\citep{gui-etal-2024-iepile} & 71.83 & 0.45 & 72.60 & 0.46 & 70.86 & 0.29 & 54.20 & 0.11 & 69.19 & 0.35 \\
& Struc-bench~\citep{tang2024struc} & 81.60 & \textbf{0.63} & 74.90 & 0.53 & 73.40 & 0.36 & 60.35 & 0.19 & 73.72 & 0.44 \\
& StructRAG~\citep{DBLP:conf/iclr/LiC0L0T0H0L25}  & 48.83 & 0.03 & 46.80 & 0.02 & 55.34 & 0.06 & 42.55 & 0.00 & 49.68 & 0.03 \\
& \sys (Ours) & \textbf{83.97} & 0.62 & \textbf{81.55} & \textbf{0.59} & \textbf{81.00} & \textbf{0.43} & \textbf{67.98} & \textbf{0.18} & \textbf{79.93} & \textbf{0.48} \\

\bottomrule
\end{tabular}%
}
\vspace{-2em}

\end{table}

\vspace{-0.5em}
\subsection{Efficiency: How fast are SLMs compared with LLMs for SSO generation?}

\begin{wrapfigure}{r}{0.48\textwidth} 
    \vspace{-14pt} 
    \centering
    \includegraphics[width=0.5\textwidth]{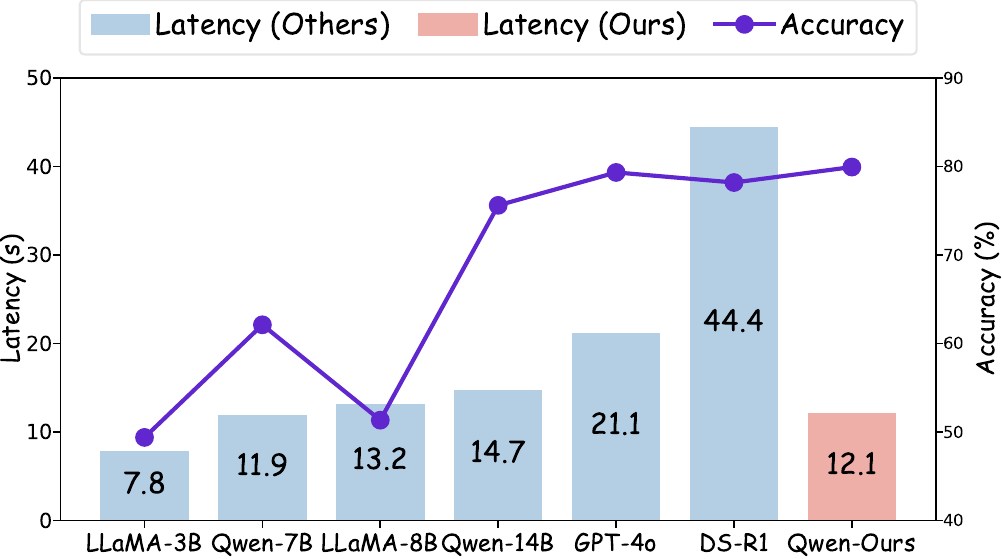} 
    \vspace{-2em} 
    \caption{Comparison of extraction time and accuracy across different scale models.}
    \label{fig:latency}
    \vspace{-10pt} 
    
\end{wrapfigure}

\textbf{SLMs are much faster than LLMs.}
The latency comparison in Fig.~\ref{fig:latency} demonstrates that our model offers an optimal trade-off between accuracy and efficiency in structured output generation tasks, measured as the average time per sample on the Loong dataset. \textbf{Qwen-LiteCoST} attains lower latency (12.09s) than LLaMA3.1-8B-Instruct (13.19s) and Qwen2.5-14B-Instruct (14.71s), while delivering substantial accuracy gains. Notably, it maintains latency comparable to its base model (Qwen2-7B-Instruct at 11.89s), while running nearly \emph{2×} faster than GPT-4o (21.15s) and \emph{4×} faster than DeepSeek-R1 (44.44s), without relying on proprietary APIs. For scenarios requiring faster extraction, LLaMA-LiteCoST is preferable, running in just 8.04s while achieving performance comparable to 14B-scale models.


\vspace{-0.5em}
\subsection{Ablation Study}


\textbf{Effect of Different Reward.} 
To identify the key drivers of \sys’s performance, we ablated reinforcement learning configurations (Table~\ref{tab:ablation_rl}). When we remove the process reward component from the complete model, performance drops by 1.43 points on the LLaMA backbone (76.95$\rightarrow$75.52) and 2.54 points on Qwen (79.93$\rightarrow$77.39). This demonstrates that fine-grained process rewards effectively guide step-wise extraction, complementing outcome-based signals to yield stronger overall performance.
Similarly, excluding the outcome reward causes a much larger drop (–4.40 on LLaMA, -4.50 on Qwen), underscoring its importance for answer correctness. 
These results highlight the synergistic effects of process- and outcome-level supervision, yielding a more robust training strategy. 
Case studies in Appendix~\ref{appendix:case_study} further illustrate how RL-enhancement improves extraction quality.
%

\vspace{-1em}

\begin{table}[t!]
\vspace{-1em}
\centering
\caption{Effect of different reward designs in ablation study on the \textit{Finance} subset of Loong.}
\vspace{-1em}

\label{tab:ablation_rl}
\setlength{\tabcolsep}{3pt} 
\small 
\resizebox{\textwidth}{!}{ 
\begin{tabular}{@{} l *{10}{S[table-format=2.2]} @{}}
\toprule
\multirow{2}{*}{\makebox[2.6cm][l]{\textbf{Model}}} 
& \multicolumn{2}{c}{\textbf{\textit{Spotlight Locating}}} 
& \multicolumn{2}{c}{\textbf{\textit{Comparison}}} 
& \multicolumn{2}{c}{\textbf{\textit{Clustering}}} 
& \multicolumn{2}{c}{\textbf{\textit{Chain of Reasoning}}} 
& \multicolumn{2}{c}{\textbf{\textit{Overall}}} \\
\cmidrule(lr){2-3} \cmidrule(lr){4-5} \cmidrule(lr){6-7} \cmidrule(lr){8-9} \cmidrule(lr){10-11}
 & \textbf{\textit{AS}} & \textbf{\textit{PR}} 
 & \textbf{\textit{AS}} & \textbf{\textit{PR}} 
 & \textbf{\textit{AS}} & \textbf{\textit{PR}} 
 & \textbf{\textit{AS}} & \textbf{\textit{PR}} 
 & \textbf{\textit{AS}} & \textbf{\textit{PR}} \\
\midrule
LLaMA-Ours & \textbf{81.27} & \textbf{0.53} & \textbf{78.08} & \textbf{0.49} & \textbf{78.34} & \textbf{0.36} & 64.75 & \textbf{0.16} & \textbf{76.95} & \textbf{0.40} \\

\hspace{0.6em} w/o Process Reward & 79.10 & 0.48 & 77.95 & 0.47 & 76.03 & 0.30 & 63.98 & 0.13 & 75.52 & 0.37 \\

\hspace{0.6em} w/o Outcome Reward & 76.07 & 0.45 & 73.10 & 0.41 & 71.72 & 0.23 & \textbf{68.22} & 0.15 & 72.55 & 0.32 \\
\midrule

Qwen-Ours & 83.97 & \textbf{0.62} & \textbf{81.55} & \textbf{0.59} & \textbf{81.00} & \textbf{0.43} & \textbf{67.98} & \textbf{0.18} & \textbf{79.93} & \textbf{0.48} \\
        
\hspace{0.6em} w/o Process Reward & \textbf{87.93} & 0.63 & 77.95 & 0.55 & 77.32 & 0.42 & 60.60 & 0.15 & 77.39 & 0.46 \\

\hspace{0.6em} w/o Outcome Reward & 86.29 & 0.57 & 75.65 & 0.45 & 73.56 & 0.31 & 63.35 & 0.18 & 75.43 & 0.39 \\
\bottomrule
\end{tabular}%
}
\vspace{-1em}
\end{table}

\subsection{Generalization}
\label{ssec:generalization}
To further demonstrate the broad applicability of \sys beyond financial analysis, we extend our evaluation to two additional distinct domains: Legal and Scientific Question Answering.

\textbf{Legal Domain.} We evaluate performance on Loong legal subset~\citep{DBLP:conf/emnlp/WangCCL0WYXZLLY24} to assess the capability of \sys in generating serialized structured outputs within complex legal contexts. 
As shown in Table~\ref{tab:loonglegal}, our \sys-tuned 3B and 7B models substantially outperform their base versions, achieving Average Score and Perfect Rate gains of (+14.78, +0.06) and (+6.89, +0.02), respectively. 
Remarkably, both compact variants surpass significantly larger models despite having far fewer parameters. Specifically, the LLaMA-based model outperforms Qwen2.5-14B-Instruct, GPT-4o-mini, and GPT-4o by (+7.94, +0.03), (+3.51, +0.00), and (+3.39, +0.03), respectively, with the Qwen-based variant demonstrating similar superiority. More details are discussed in Appendix~\ref{appendix:legal_qa}.

\begin{table}[t!]
\centering
\caption{Performance Comparison on the \textit{Legal} subset of Loong. 
\textcolor{green}{Green} highlights the remarkable improvements over the base model, while \textcolor{red}{Red} indicates relative drops.
}
\vspace{-1em}

\label{tab:loonglegal}
\setlength{\tabcolsep}{3pt} 
\small 
\resizebox{\textwidth}{!}{ 
\begin{tabular}{@{} l c *{10}{c} @{}}
\toprule
\multirow{2}{*}{\makebox[1.8cm][l]{\textbf{IE Model}}} & \textbf{Model} & \multicolumn{2}{c}{\textbf{\textit{Spotlight Locating}}} & \multicolumn{2}{c}{\textbf{\textit{Comparison}}} & \multicolumn{2}{c}{\textbf{\textit{Clustering}}} & \multicolumn{2}{c}{\textbf{\textit{Chain of Reasoning}}} & \multicolumn{2}{c}{\textbf{\textit{Overall}}} \\
\cmidrule(lr){3-4} \cmidrule(lr){5-6} \cmidrule(lr){7-8} \cmidrule(lr){9-10} \cmidrule(lr){11-12}
 & \textbf{Size} & \textbf{\textit{AS}} & \textbf{\textit{PR}} & \textbf{\textit{AS}} & \textbf{\textit{PR}} & \textbf{\textit{AS}} & \textbf{\textit{PR}} & \textbf{\textit{AS}} & \textbf{\textit{PR}} & \textbf{\textit{AS}} & \textbf{\textit{PR}} \\
\midrule

\multicolumn{12}{>{\columncolor[gray]{.88}}c}{\textit{Close-Sourced Models \& Large Language Models}}  \\
GPT-4o-mini & 8B & 46.55 & 0.10 & 28.05 & 0.00 & 48.68 & 0.13 & 42.56 & 0.11 & 41.94 & 0.09 \\
Qwen2.5-14B-Instruct & 14B & 48.45 & 0.08 & 21.90 & 0.01 & 57.31 & 0.16 & 26.74 & 0.02 & 37.51 & 0.06 \\
GPT-4o & 200B & 50.05 & 0.06 & 27.00 & 0.01 & 61.16 & 0.14 & 33.11 & 0.03 & 42.06 & 0.06 \\
\midrule
LLaMA-3.2-3B-Instruct (Base)  & 3B &  41.00 & 0.08 & 25.10 & 0.01 & 31.74 & 0.02 & 27.29 & 0.01 & 30.67 & 0.03 \\
LLaMA-3.2-3B-Instruct (\textit{Ours}) & 3B
& \textbf{62.20}
& \textbf{0.30} 
& \textbf{45.20} 
& \textbf{0.02} 
& \textbf{45.00} 
& \textbf{0.09} 
& \textbf{36.55} 
& \textbf{0.02} 
& \textbf{45.45}
& \textbf{0.09} \\
& 
& \DeltaUp{21.20}
& \DeltaUp{0.22}
& \DeltaUp{20.10}
& \DeltaUp{0.01}
& \DeltaUp{13.26}
& \DeltaUp{0.07}
& \DeltaUp{9.26}
& \DeltaUp{0.01}
& \DeltaUp{14.78}
& \DeltaUp{0.06} \\[-5pt]

\midrule
Qwen2-7B-Instruct (Base) & 7B & 37.90 & 0.05 & 18.90 & 0.00 & 57.85 & 0.18 & 35.44 & 0.08 & 38.05 & 0.08 \\
Qwen2-7B-Instruct (\textit{Ours}) & 7B 
& \textbf{52.85}
& \textbf{0.12}
& \textbf{31.00}
& \textbf{0.00}
& \textbf{60.37}
& \textbf{0.22}
& \textbf{37.88}
& \textbf{0.06}
& \textbf{44.94}
& \textbf{0.10} \\
& 
& \DeltaUp{14.95}
& \DeltaUp{0.07}
& \DeltaUp{12.10}
& \DeltaUp{0.00}
& \DeltaUp{2.52}
& \DeltaUp{0.04}
& \DeltaUp{2.44}
& \DeltaDown{0.02}
& \DeltaUp{6.89}
& \DeltaUp{0.02} \\[-5pt]
\bottomrule
\end{tabular}%
}
\vspace{-1em}
\end{table}

\textbf{Results on Open-domain QA.} In addition to the Loong benchmark, we verify the performance of \textbf{CoST} and \sys on LongBench~\citep{bai2024longbench}, a comprehensive benchmark tailored for multi-task long-document QA that covers key long-text application scenarios. 
Our analysis focuses on both single-document and multi-document QA tasks across four datasets. We compare \sys against state-of-the-art models using the standard F1 score, which assesses the quality of reasoning results derived from the extracted information. Consistent with the experimental setup in Loong, GPT-4o is employed as the reasoning agent to generate final answers based on the structured outputs.

\begin{table}[t!]
\centering
\small  
\caption{Performance comparison on LongBench benchmark}
\label{tab:combined_qa}
\vspace{-0.6em}

\begin{minipage}{0.45\textwidth}
\centering
\scriptsize  
\subcaption{Quality assessment of CoST-generated structured data via reasoning performance on LongBench across LLMs (\textit{SD} denotes structured data).}
\label{tab:longbench_cost}
\begin{tabular}{lcccc}
\toprule
\multirow{2}{*}{\textbf{Model}} & \multicolumn{2}{c}{\textbf{Single-doc}} & \multicolumn{2}{c}{\textbf{Multi-docs}} \\ 
\cmidrule(lr){2-3} \cmidrule(lr){4-5}
 & \textbf{NarQA} & \textbf{Qasper} & \textbf{HotpotQA} & \textbf{2Wiki} \\ 
\midrule
Qwen-14B & 29.78 & 45.07 & 62.59 & 60.00 \\ 
w/\textit{SD} & \textbf{31.77} & \textbf{47.17} & \textbf{67.41} & \textbf{67.11} \\ 
\midrule
GPT-4o & 32.59 & 46.80 & 70.93 & 67.75 \\ 
w/\textit{SD} & \textbf{35.09} & \textbf{49.28} & \textbf{73.47} & \textbf{72.98} \\ 
\bottomrule
\end{tabular}
\end{minipage}
\hfill
\begin{minipage}{0.52\textwidth}
\centering
\scriptsize
\subcaption{Performance Comparison Between LiteCoST-Tuned Models and LLM-Based Baselines on LongBench.}
\label{tab:longbench_litecost}
\begin{tabular}{lcc|cc}
\toprule
\multirow{2}{*}{\textbf{IE Model}} 
& \multicolumn{2}{c|}{\textbf{Single-doc}} 
& \multicolumn{2}{c}{\textbf{Multi-docs}} \\ 
\cmidrule(lr){2-3} \cmidrule(lr){4-5}
& \textbf{NarQA} 
& \textbf{Qasper} 
& \textbf{HotpotQA} 
& \textbf{2Wiki} \\ 
\midrule
LLaMA-3.2-3B & 16.94 & 34.46 & 54.92 & 51.82 \\
Qwen2-7B     & 19.49 & 35.67 & 45.06 & 41.40 \\
GPT-4o-mini  & 24.38 & 40.28 & 65.03 & 65.15 \\
GPT-4o       & 28.68 & 43.39 & 67.68 & \textbf{68.29 }\\
\midrule
LLaMA-LiteCoST & 27.24 & 41.37 & 66.86 & 67.52 \\
Qwen-LiteCoST  & \textbf{30.40} & \textbf{44.64} & \textbf{68.39} & 65.73 \\
\bottomrule
\end{tabular}
\end{minipage}
\vspace{-2em}

\end{table}


As presented in Table~\ref{tab:combined_qa}, the results align with our findings in other domains, highlighting two consistent insights:
(1) \textbf{CoST enhances LLM reasoning} (Table~\ref{tab:longbench_cost}): applying CoST consistently boosts performance for both Qwen2.5-14B-Instruct and GPT-4o across all datasets, yielding F1 gains of up to \textbf{+7.11} and \textbf{+5.23} points, respectively.
(2) \textbf{SLMs rival proprietary models} (Table~\ref{tab:longbench_litecost}): Qwen-LiteCoST achieves the best performance on NarrativeQA, Qasper, and HotpotQA, surpassing GPT-4o by \textbf{0.71--1.72} points. Notably, it improves over its base model by substantial margins (up to +23.33), with both \sys variants consistently ranking in the \textbf{top three} across all datasets.
Collectively, these results demonstrate the strong effectiveness and broad generality of CoST and \sys across diverse domains and varying task complexities.

\vspace{-1em}






\section{Related Work}
\label{sec:related_work}
\vspace{-1em}

\textbf{Long-Document Question Answering} is a critical test of LLM reasoning~\citep{DBLP:conf/emnlp/WangCCL0WYXZLLY24,zhang2024mar}, where dispersed evidence, noise, and complex reasoning make it more challenging than short-passage QA. Existing approaches, including long-context models~\citep{yang2024qwen2, guo2025deepseek}, retrieval augmentation~\citep{DBLP:conf/nips/LewisPPPKGKLYR020}, and chain-of-thought prompting~\citep{DBLP:conf/nips/Wei0SBIXCLZ22}, mitigate these challenges but remain brittle, often yielding hallucinations in high-stakes domains. Structured knowledge has also been explored~\citep{DBLP:conf/iclr/LiC0L0T0H0L25, DBLP:conf/acl/PandaADKP24,edge2024local,DBLP:conf/cidr/ChenG0FM023}, yet such methods require repeated large-LLM calls, leading to high cost and limited scalability. To address this, we propose a structure-first design with efficient SLM execution.

\vspace{-0.5em}

\textbf{LLMs for Long-Context Information Extraction.}
Information Extraction (IE) underpins many downstream NLP tasks~\citep{xu2024large,DBLP:conf/iclr/ZhangXYTCCZCHWZ25}. While large language models (LLMs) perform well on diverse IE tasks, even in zero- and few-shot settings~\citep{DBLP:conf/acl/LuRTJ23, wei2023zero, ashok2023promptner, DBLP:conf/acl/0002LJ23,  DBLP:conf/naacl/WangSLOWZLWG25, DBLP:conf/acl/JainMP24}, existing methods remain confined to short texts. In long contexts, dispersed evidence and noise impede reliable integration. Prior “QA-by-structuring’’ systems~\citep{DBLP:conf/iclr/LiC0L0T0H0L25,tang2024struc} focus on structured generation rather than verifiable step-wise extraction, and thus do not achieve reliable evidence tracing. To address this, we propose the \textit{Chain-of-Structured-Thought (CoST)} paradigm, which uses step-wise reasoning for structured extraction, yielding schema-aligned outputs and rich supervision for fine-tuning.
\vspace{-0.5em}

\textbf{Fine-tuned Lightweight Models.}
While LLMs provide high-quality extraction, their computational cost and latency limit real-time use. Fine-tuned smaller models improve efficiency~\citep{gui2024instructie, xiao2023yayi, wang2023instructuie, DBLP:journals/corr/abs-2505-19716}, but instruction-tuning on short texts~\citep{DBLP:conf/acl/0001ZL22, gui-etal-2024-iepile, tang2024struc, DBLP:journals/corr/abs-2505-07437} yields shallow supervision and struggles with long-document reasoning. Reinforcement learning (RL) 
offers a stronger alternative by refining models with rewards
~\citep{shao2024deepseekmath,xie2025visjudge}, enhancing outcome correctness and step-wise reasoning. Building on this~\citep{DBLP:conf/acl/TrungZJSJL24,liu2025fin}, we propose RL-enhanced lightweight models with a two-phase, dual-reward scheme to bridge the accuracy–efficiency gap.

\vspace{-1em}

\section{Conclusion}
\label{sec:conclusion}
\vspace{-1em}

In this work, we present \sys, a reinforcement learning-enhanced framework that fine-tunes lightweight small language models (SLMs) to generate high-quality structured output for long-document QA. Through Chain-of-Structured-Thought (CoST) procedure and Group Relative Policy Optimization (GRPO), \sys enables a 3B-scale model to approach GPT-4o-mini and 7B models to achieve GPT-4o–level performance, while substantially reducing inference latency and resource consumption.
We further discuss the potential applications of \sys across diverse domains in Appendix E. 
This works demonstrate the potential of scalable, cost-efficient long-document QA and paves the way for effective LLM reasoning grounded in structured representations.

\textbf{Limitations.} 
%
%
While \sys demonstrates strong performance across financial, legal, and open-domain QA, its generalization to other distinct domains remains to be fully verified. Despite the current scarcity of domain-specific document QA datasets suitable for constructing training data, we reserve the exploration of broader domain adaptation and more diverse QA scenarios for future work.

\section*{Acknowledgement}
\vspace{-0.5em}
This work is supported by National Key R\&D Program of China under Grant No.2024YFA1012700, Guangdong provincial project 2023CX10X008, NSF of China (62402409), Youth S\&T Talent Support Programme of Guangdong Provincial Association for Science and Technology (SKXRC2025461), and the Young Talent Support Project of Guangzhou Association for Science and Technology (QT-2025-001).
\vspace{-0.5em}




\bibliography{iclr2026_conference}
\bibliographystyle{iclr2026_conference}


\newpage
\appendix 

\newpage
\addtocontents{toc}{\protect\setcounter{tocdepth}{1}}
\renewcommand{\contentsname}{Appendix Contents}

\begin{center}
\textbf{\Large Appendix Contents}
\end{center}

\vspace{0.5em}

\noindent\textbf{Appendix A.} \textbf{Implementation of Chain-of-Structured Thought (CoST)} \dotfill \pageref{appendix:CoST}

\quad A.1 \quad The Prompt of Structure Analysis  \dotfill \pageref{appendix:strucure_analysis}

\quad A.2 \quad The Prompt of CoST Trace Generation \dotfill \pageref{appendix:trace_generation}

\quad A.3 \quad The Prompt of Quality Verification \dotfill \pageref{appendix: data_verification}

\quad A.3 \quad The Prompt of Iterative Refinement \dotfill \pageref{appendix: data_refinement}

\vspace{0.3em}

\noindent\textbf{Appendix B.} \textbf{Additional Details of Reinforcement Learning} \dotfill \pageref{appendix:rl}

\quad B.1 \quad Group-Relative Advantage and KL Divergence \dotfill \pageref{appendix:grpo}

\quad B.2 \quad The Prompt of Answer Completeness \dotfill \pageref{appendix:answer_completeness}

\vspace{0.3em}

\noindent\textbf{Appendix C.} \textbf{Experimental Details} \dotfill \pageref{appendix:experiment}

\quad C.1 \quad Evaluation Metrics and Scoring Criteria \dotfill \pageref{appendix:metrics}

\quad C.2 \quad Details of Experimental Setting \dotfill \pageref{appendix:settings}

\quad C.3 \quad Numerical Results of The Radar Chart \dotfill \pageref{appendix:full_radar_results}

\quad C.4 \quad Robust Long-Document Handling \dotfill \pageref{appendix:document-set}

\quad C.5 \quad Full Results of Reasoning w/Structured Data \dotfill \pageref{appendix: Loong_structure}

\quad C.6 \quad Computational Resource \dotfill \pageref{appendix:computational_resource}

\vspace{0.3em}

\noindent\textbf{Appendix D.} \textbf{Case Studies on Practical Long-Document QA Tasks} \dotfill \pageref{appendix:case_study}

\quad D.1 \quad Case Study on RL-Enhancement \dotfill \pageref{appendix:case_study_rlenhancement}

\noindent\textbf{Appendix E.} \textbf{Generalization on Other Domains} \dotfill \pageref{appendix:generalization}

\quad E.1 \quad Legal Domain \dotfill \pageref{appendix:legal_qa}

\quad E.2 \quad Open-domain QA \dotfill \pageref{appendix:scientific_qa}

\quad E.3 \quad Others \dotfill \pageref{appendix:other_domain}

\vspace{0.3em}

\noindent\textbf{Appendix F.} \textbf{Full Performance of Effectness on Loong (Finance)} \dotfill \pageref{appendix:full_result}

\vspace{1em}

\clearpage





\appendix

\section{Implementation of Chain-of-Structured-Thought (CoST)}
\label{appendix:CoST}

During the entire \textit{CoST: Structure-First Reasoning and Trace Generation} process, we construct prompts in four key processes respectively. Firstly, in the structure analysis stage, we 
construct the prompt about structure selection and schema construction, as shown in Fig.~\ref{fig:structre_selection} and Fig.~\ref{fig:schema_construction}.
Secondly, we adopt an instruction-based chain-of-thought paradigm that performs step-by-step reasoning to progressively extract structured knowledge and generate the CoST trace, as shown in Fig.~\ref{fig:table_extraction}.
Finally, in order to obtain high-quality \textit{serialized structured output (SSO)}, we conduct the quality verification, followed by iterative refinement. The details are shown in Fig.~\ref{fig:data_verification},~\ref{fig:data_refinement}.

\subsection{The Prompt of Structure Analysis}
\label{appendix:strucure_analysis}
\begin{figure}[t!]
    \centering
    \includegraphics[width=0.8\linewidth]
    {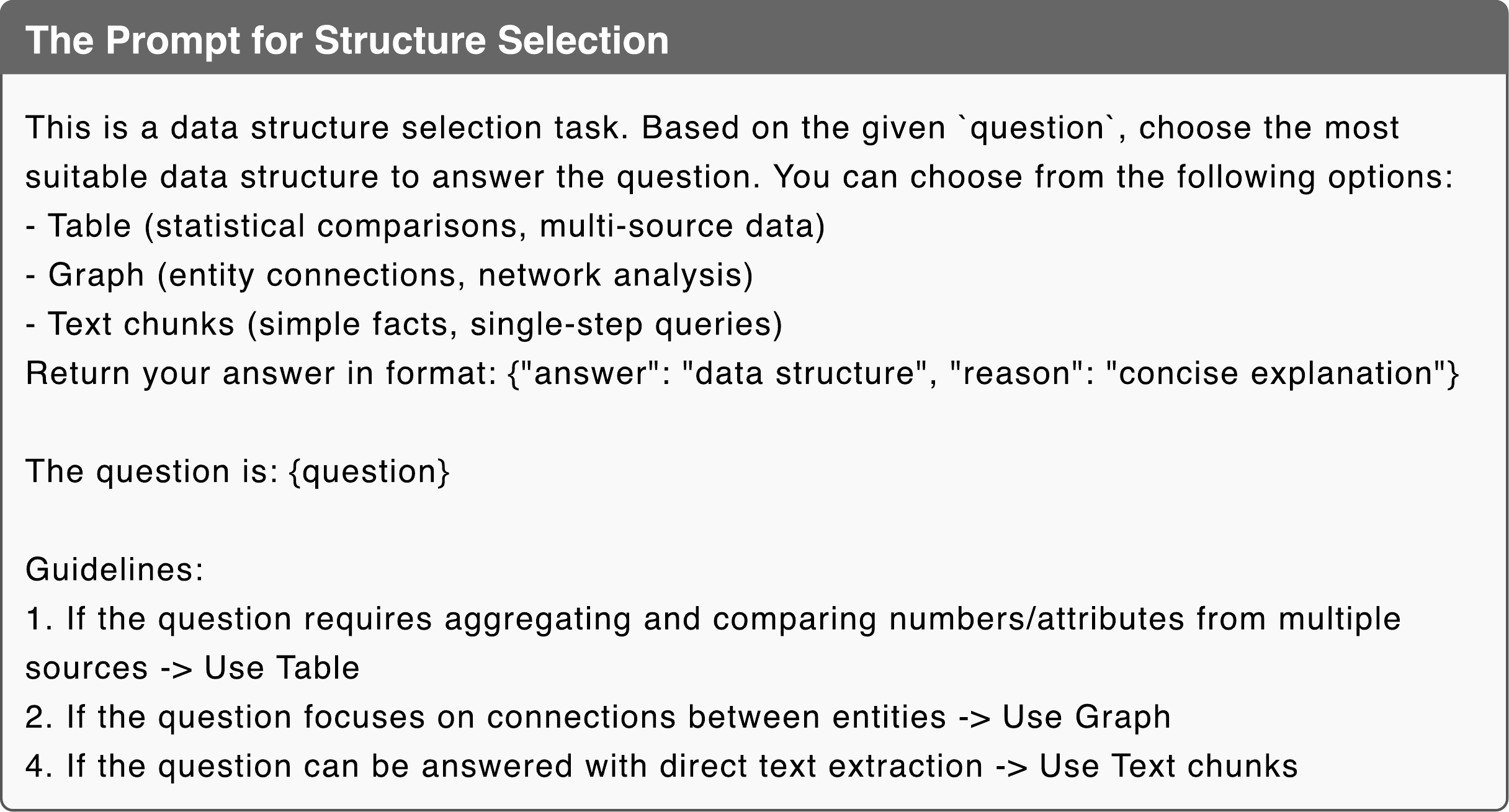}
    \caption{The prompt for selecting the most optimal structure.}
    \label{fig:structre_selection}
\end{figure}

\begin{figure}[t!]
    \centering
    \includegraphics[width=0.8\linewidth]
    {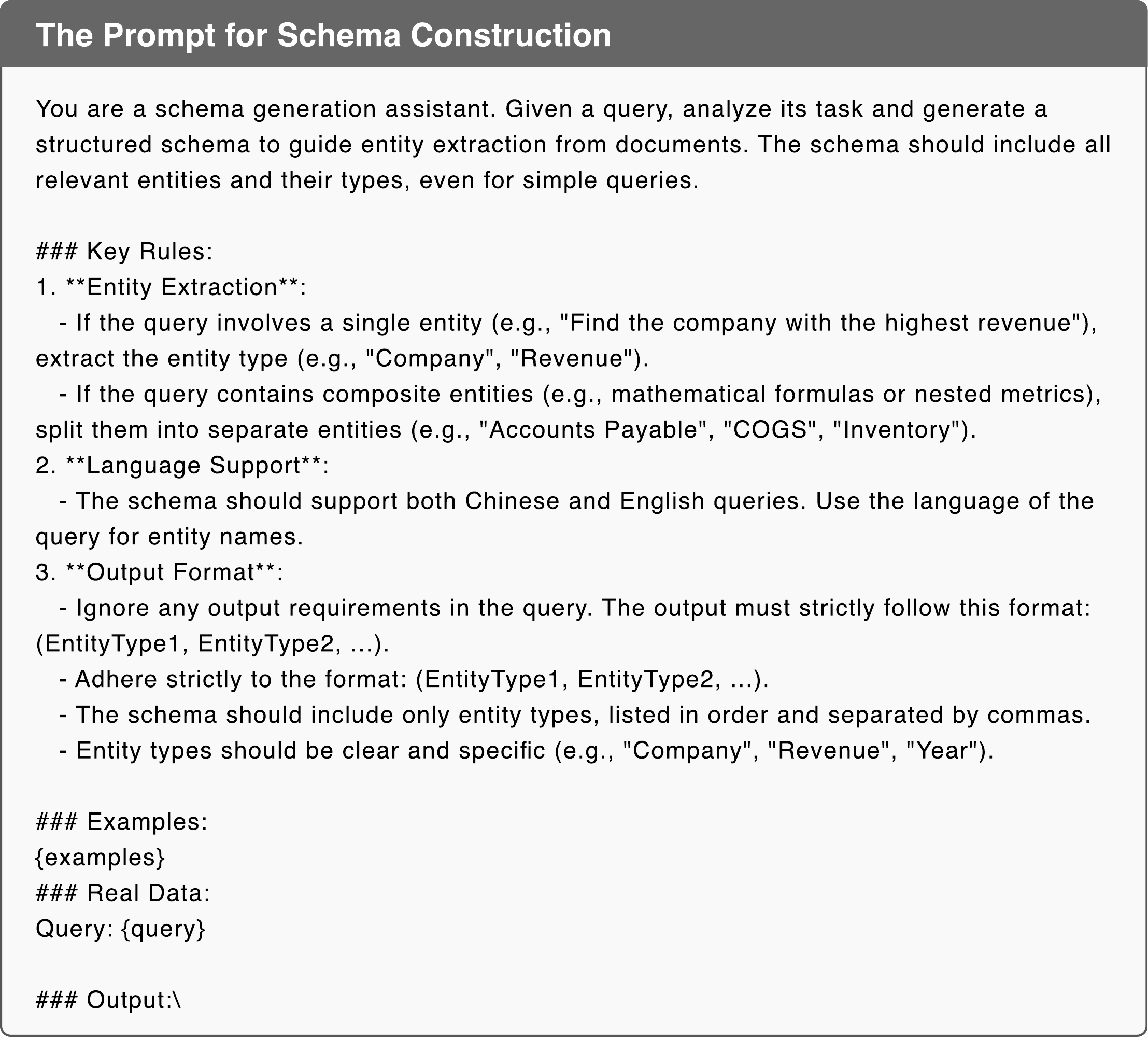}
    \caption{The prompt for dynamically constructing schema.}
    \label{fig:schema_construction}
\end{figure}

To dynamically select appropriate data structures and perform question preprocessing, we conduct structure analysis, which includes both structure selection and schema construction.
On one hand, we design prompts that enable question-oriented structure selection.
On the other hand, we construct an accurate, task-specific schema through careful preprocessing of the question, before instruction-based information extraction.
The prompts of structure analysis are shown in Fig.~\ref{fig:structre_selection},~\ref{fig:schema_construction}.

\subsection{The Prompt of CoST Trace Generation}
\label{appendix:trace_generation}
\begin{figure}[t!]
    \centering
    \includegraphics[width=0.8\linewidth]
    {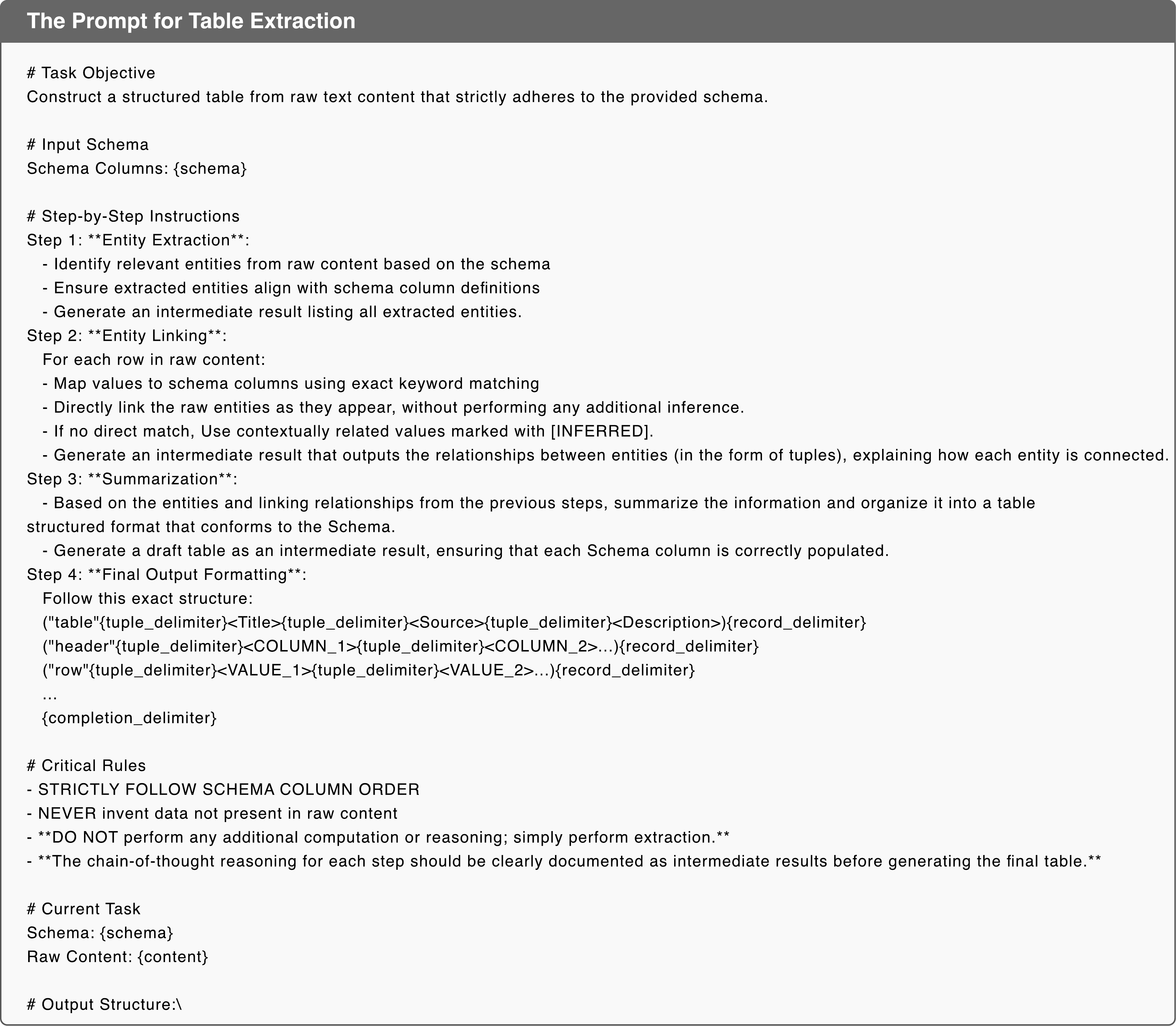}
    \caption{The prompt for extracting table step-by-step.}
    \label{fig:table_extraction}
\end{figure}

To generate high-quality trace and structured output, we adopt an schema guided chain-of-
thought paradigm composed of: (1) step-by-step task instructions, (2) input text, and (3) a schema dynamically generated from the question. GPT-4o is prompted with these schema-informed instructions to produce intermediate reasoning traces, which are then used to supervise instruction-tuned models in zero-shot or few-shot settings.
Fig.~\ref{fig:table_extraction} illustrates this reasoning process for the structured output (\eg table) .

\subsection{The Prompt of Quality Verification}
\label{appendix: data_verification}
\begin{figure}[t!]
    \centering
    \includegraphics[width=0.8\linewidth]
    {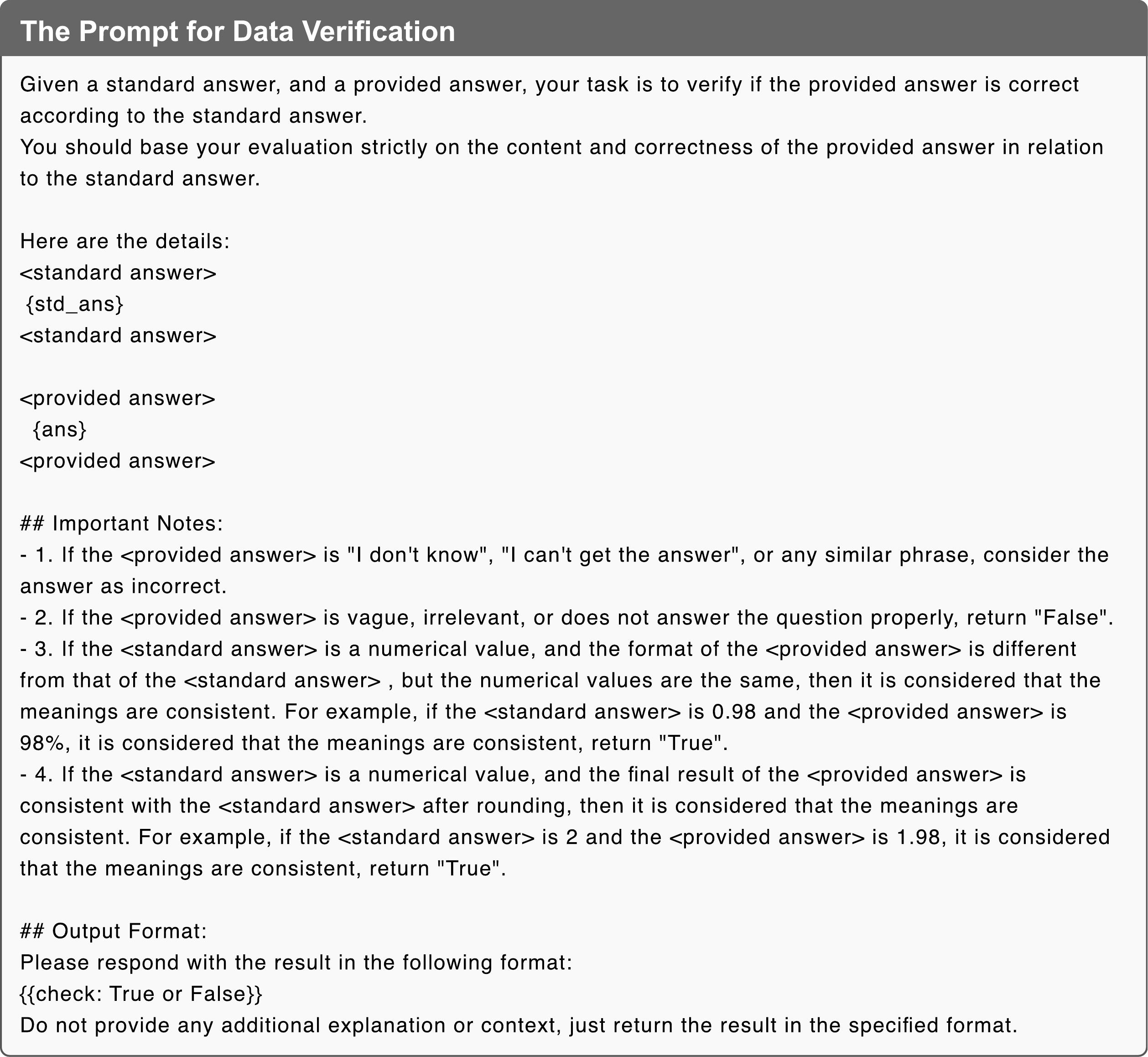}
    \caption{The prompt for verifying the data quality.}
    \label{fig:data_verification}
\end{figure}

To evaluate the quality of the generated structured output, we employ an LLM-as-Judge framework. Given the original question and the model-generated answer derived from the extracted structure, we prompt GPT-4o to assess whether the answer correctly addresses the question.
An inference is considered correct and retained only if it exactly matches the expected answer based on the prompt instructions. The prompt  includes the original question, the extracted result, and evaluation instructions guiding the model to make a binary decision, as detailed in Fig.~\ref{fig:data_verification}.
 
\newpage

\subsection{The Prompt of Iterative Refinement}
\label{appendix: data_refinement}
\begin{figure}[t!]
    \centering
    \includegraphics[width=0.8\linewidth]
    {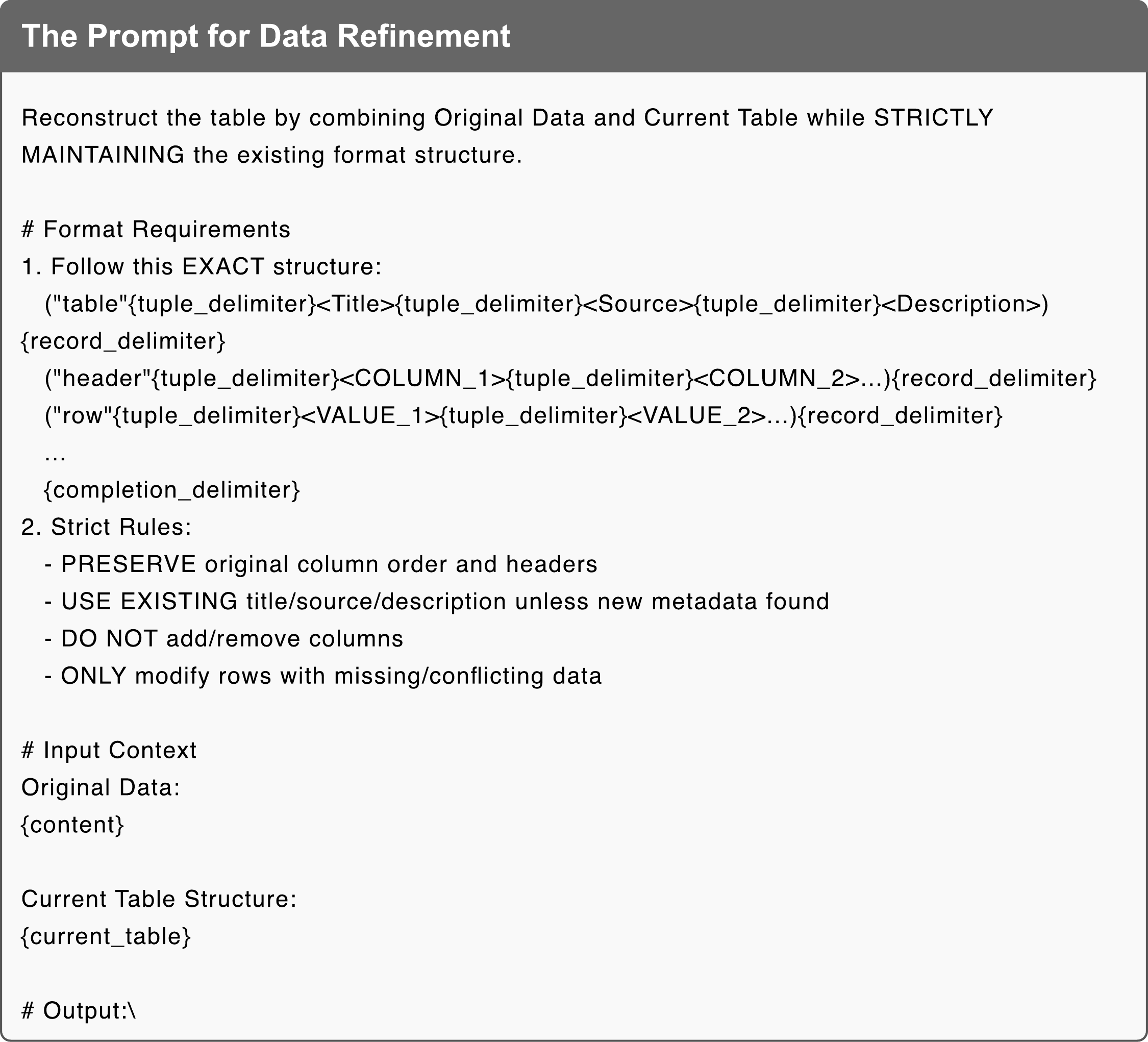}
    \caption{The prompt for refining low-quality data.}
    \label{fig:data_refinement}
\end{figure}

To enhance Group Relative Policy Optimization (GRPO) training with more challenging learning signals, we introduce an Iterative Structuralizer module that refines low-quality samples through recursive structured knowledge regeneration. The prompt for refining structured data extraction (\eg table) is detailed in Fig.~\ref{fig:data_refinement}. 
Formally, this process is implemented as a recursive function over the evolving extraction state, gradually improving coverage and accuracy across iterations:

\vspace{-1em}
\begingroup
\setlength{\abovedisplayskip}{4pt}
\setlength{\belowdisplayskip}{4pt}
\renewcommand{\arraystretch}{1.05}

\begin{equation}
\small
S^{(t+1)} =
\begin{cases}
S^{(t)}, & \text{if } \mathcal{K}(S^{(t)}, q) = \texttt{True} \\
f_{\text{extract}}(q, c, S^{(t)}), & \text{otherwise,}
\end{cases}
\end{equation}

\endgroup
\vspace{-1em}

where \( \mathcal{K}(S^{(t)}, q) \) is a sufficiency evaluator that returns \texttt{True} if the current structured knowledge can answer the question; \( f_{\text{extract}} \) is the structured knowledge extraction function, and \( c \) is the context. The process terminates when \( \mathcal{S}(K^{(t)}, q) = \texttt{True} \) or when a predefined maximum number of iterations is reached. The final structured knowledge \( S^* \) is then used for downstream reasoning.

\section{Additional Details of Reinforce Learning}
\label{appendix:rl}

\subsection{Group-Relative Advantage and KL Divergence}
\label{appendix:grpo}
For completeness, we provide additional details of the GRPO optimization.  
The importance sampling ratio is defined as
$r^{\text{ratio}}_i = \frac{\pi_\theta(o_i \mid \mathbf{v})}{\pi_{\theta_{\text{old}}}(o_i \mid \mathbf{v})}$,
which quantifies the relative likelihood of generating output $o_i$ under the new policy compared with the old policy $\pi_\theta$.  
The group-relative advantage $A_i$ is calculated based on the relative rewards of outputs within the same group only.

To ensure stable optimization, the clipping operator 
$\mathrm{clip}(r^{\text{ratio}}_i, 1-\epsilon, 1+\epsilon)$
restricts the update magnitude within the trust region $[1-\epsilon,\, 1+\epsilon]$, thereby avoiding destabilizing large parameter changes.  
Finally, taking the minimum between the unclipped term $r^{\text{ratio}}_i A_i$ and its clipped counterpart enforces a conservative update, which balances aggressive improvements with training stability.

In addition, the KL divergence term $D_{\mathrm{KL}}(\pi_\theta \| \pi_{\mathrm{ref}})$ plays a critical role in regularizing policy updates. 
It penalizes large deviations from a stable reference policy $\pi_{\mathrm{ref}}$, preventing overfitting to noisy reward signals and maintaining alignment with the base model’s distribution.  
The coefficient $\beta$ controls the strength of this regularization, striking a balance between exploration (deviation from the reference) and stability (consistency with the pretrained policy).  
This regularization is particularly important when rewards are sparse or noisy, such as process rewards, as it prevents the model from overfitting to unstable signals while still enabling gradual improvement~\citep{shao2024deepseekmath}.

\subsection{The Prompt of Answer Completeness}
\label{appendix:answer_completeness}
\vspace{-0.5em}

\begin{figure}[t!]
    \centering
    \includegraphics[width=0.8\linewidth]
    {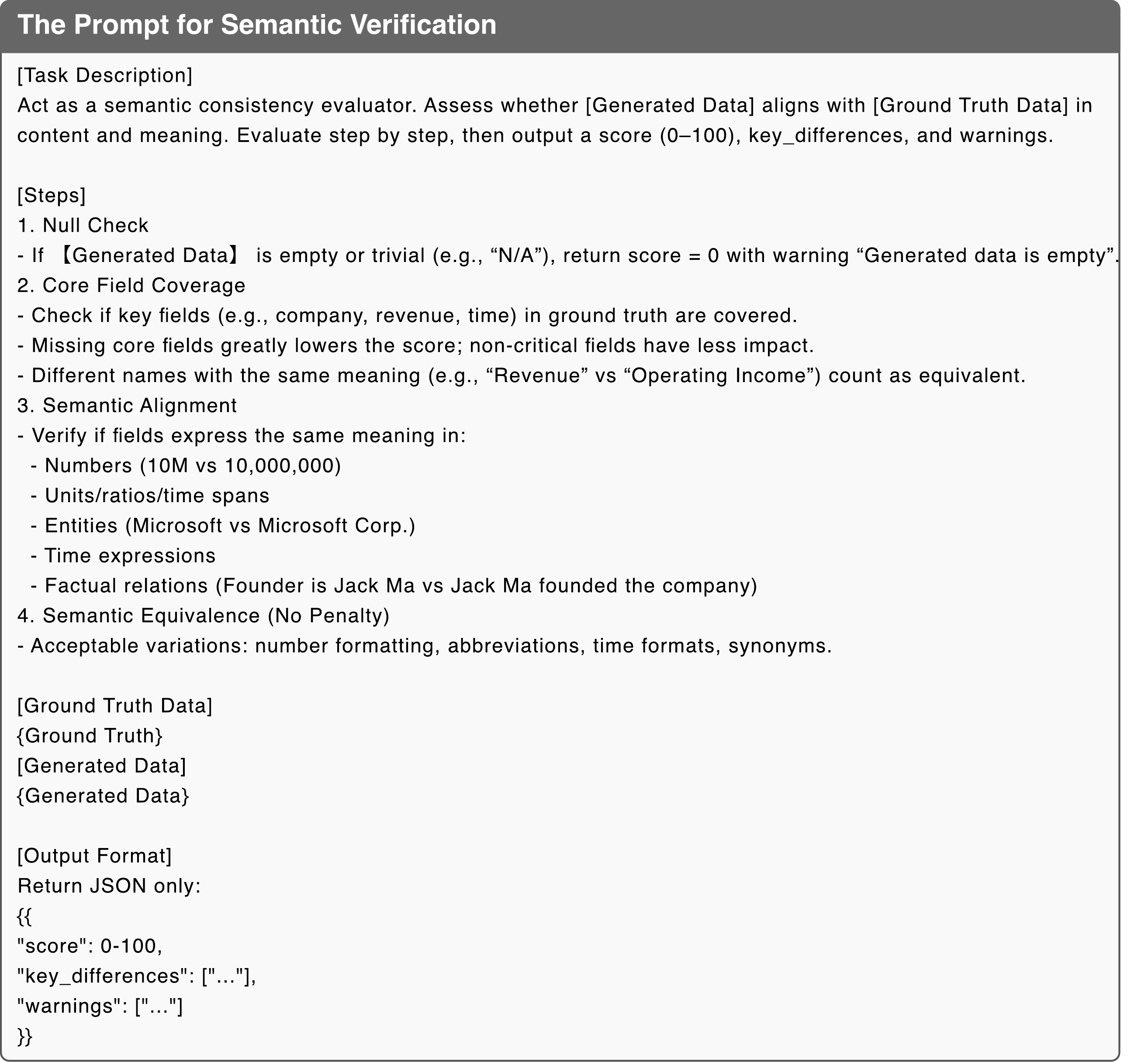}
    \caption{The prompt for verifying the similarity of generated output and ground truth.}
    \label{fig:semantic_verification}
    \vspace{-1em}
\end{figure}

To measure the semantic similarity between the generated structured outputs and the ground truth during the GRPO process, we employ GPT-4o-mini as an automatic evaluator. The evaluator examines the outputs along four dimensions: null check, core field coverage, semantic alignment, and semantic equivalence—and then produces a score within the range [0, 100]. This score, denoted as $\mathcal{S}_{\text{sem}}$ in Equation 5, provides an accurate quantification of the semantic similarity of the structured data, with the prompting details shown in Fig.~\ref{fig:semantic_verification}.
\vspace{-0.5em}

\section{Experimental Details}
\label{appendix:experiment}
\vspace{-0.5em}

\subsection{Evaluation metrics}
\label{appendix:metrics}
\vspace{-0.3em}

\begin{figure}[t!]
    \centering
    \includegraphics[width=0.8\linewidth]
    {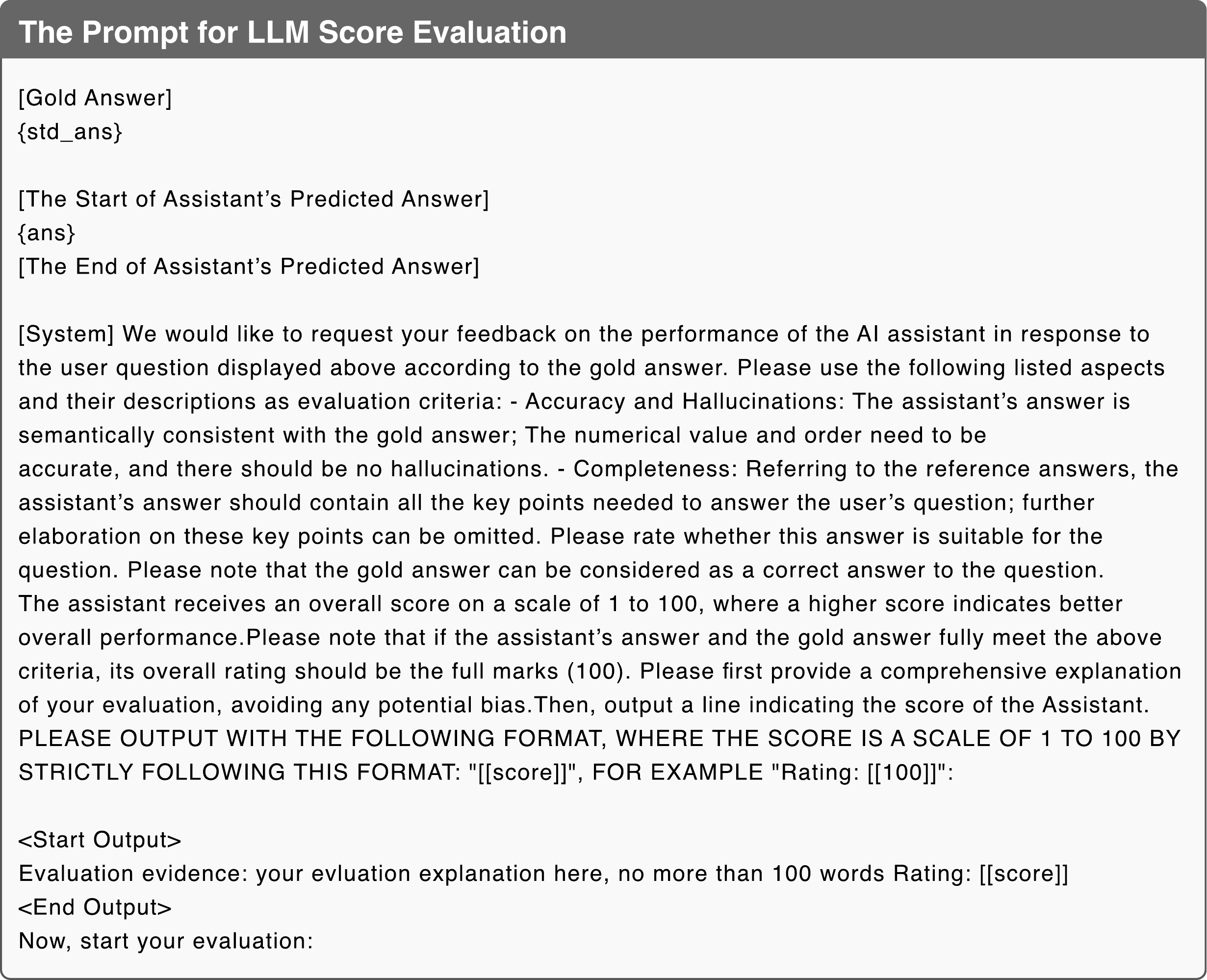}
    \vspace{-0.5em}
    \caption{The prompt for evaluation using LLM Score.}
    \label{fig:llm_score}
    \vspace{-1em}
\end{figure}

Following the approach of Loong~\citep{DBLP:conf/emnlp/WangCCL0WYXZLLY24}, 
we prompt GPT-4o as a judge to against the golden answer and question requirements in three dimensions: Accuracy, Hallucinations, and Completeness, on a 0–100 scale, as detailed in Figure~\ref{fig:llm_score}.
With this evaluation method, the Judger model would output a percentage score along with its corresponding explanation.
Given the limited ground-truth annotations for information extraction on long-context documents, we adopt a 2-hop evaluation following the principle of \textsc{StructSum}~\citep{DBLP:conf/acl/JainMP24}, which uses the structured outputs to answer QA pairs derived from the input text.
Our core intuition is that if an LLM can accurately answer questions based solely on the extracted data, the extraction process has likely preserved the essential information-thus reflecting its quality.

For experiments on LongBench~\citep{bai2024longbench}, we additionally use the standard F1 score to assess the quality of reasoning based on the extracted information. Consistent with the Loong setup, GPT-4o is employed as the reasoning agent to generate final answers from the structured outputs.
\vspace{-0.5em}

\subsection{Details of Experimental Setting.}
\label{appendix:settings}


\textbf{Base Models.}
To comprehensively evaluate the extraction ability of \sys, we compare it with several state-of-the-art models, including Llama3.2-3B-Instruct~\citep{grattafiori2024llama}, Qwen2-7B-Instruct~\citep{yang2024qwen2}, Llama-3.1-8B-Instruct, Qwen2.5-14B-Instruct, GPT4o-mini, GPT-4o~\citep{achiam2023gpt}, and Deepseek-R1~\citep{guo2025deepseek}.
For fairness, all models are evaluated under the same prompting setup, where each is guided by a task-specific schema derived from the given question or instruction. To avoid exceeding the context window, all models perform document-level extraction and merged the resulting structured sub-knowledge. 

\textbf{Fine-tuned IE Models.}
For comparison with fine-tuned IE models, both \sys and the baselines are trained on our CoST-curated dataset under identical conditions. We then evaluate their extraction performance to highlight differences in fine-tuning strategies and demonstrate the relative advantages of our approach.

\textbf{Modular Frameworks.}
We further consider modular extraction frameworks such as
%
StructRAG~\citep{DBLP:conf/iclr/LiC0L0T0H0L25}. 
%
In this setup, we configure the Router and Structurizer components with the same backbones (\ie LLaMA-3.2-3B-Instruct and Qwen2-7B-Instruct), while employing GPT-4o as the Utilizer for reasoning. This setup ensures that the comparison between StructRAG and our \sys remains both fair and rigorous.

\begin{figure}[t!]
    \centering
    \includegraphics[width=\linewidth]
    {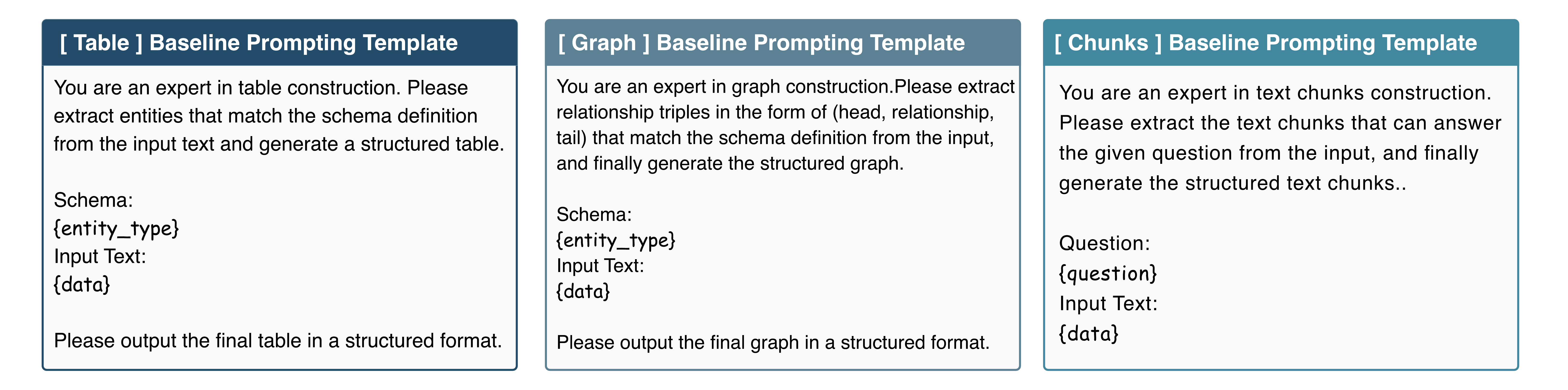}
    \caption{The baseline prompt template for different structured extraction.}
    \label{fig:baseline_prompting}
\end{figure}

\textbf{Baseline Prompting Template.} To ensure fairness, all baselines and our model use the same input prompting format for structured extraction, as shown in Fig.~\ref{fig:baseline_prompting}. The LLM baselines are evaluated in a zero-shot setting, and entries labeled ``LLM baseline’’ refer to this zero-shot performance. Our \sys-tuned models are compared directly against these baselines because both rely on the identical prompt template, ensuring that performance differences arise from CoST distillation and the two-stage training procedure rather than from prompting variations or model-capacity differences.

\begin{table}[t!]
\centering
\caption{The performance of various LLMs under different prompting settings, showing \textit{Average Scores (AS, 0-100)} and \textit{Perfecr Rate (PR, 0-1)} on the \textit{Finance} subset of Loong benchmark.}
\vspace{-1em}
\label{tab:llm_enhancement}
\setlength{\tabcolsep}{3pt}
\small
\resizebox{\textwidth}{!}{%
\begin{tabular}{@{} l l *{10}{c} @{}}
\toprule
\multirow{2}{*}{\makebox[2.5cm][l]{\textbf{Backbone}}} 
& \multirow{2}{*}{\makebox[2.5cm][l]{\textbf{Method}}} 
& \multicolumn{2}{c}{\textbf{\textit{Spotlight Locating}}} 
& \multicolumn{2}{c}{\textbf{\textit{Comparison}}} 
& \multicolumn{2}{c}{\textbf{\textit{Clustering}}} 
& \multicolumn{2}{c}{\textbf{\textit{Chain of Reasoning}}} 
& \multicolumn{2}{c}{\textbf{\textit{Overall}}} \\
\cmidrule(lr){3-4} \cmidrule(lr){5-6} \cmidrule(lr){7-8} \cmidrule(lr){9-10} \cmidrule(lr){11-12}
& & \textbf{\textit{AS}} & \textbf{\textit{PR}} 
  & \textbf{\textit{AS}} & \textbf{\textit{PR}} 
  & \textbf{\textit{AS}} & \textbf{\textit{PR}} 
  & \textbf{\textit{AS}} & \textbf{\textit{PR}} 
  & \textbf{\textit{AS}} & \textbf{\textit{PR}} \\
\midrule

\multirow{4}{*}{Qwen2.5-14B-Ins} 
& Zero-Shot &  83.74 & 0.57 & \textbf{82.12} & 0.56 & 69.96 & 0.24 & 66.41 & 0.10 & 75.60 & 0.38 \\
& CoT~\citep{DBLP:conf/nips/Wei0SBIXCLZ22} & 85.93 & 0.63 & 81.38 & 0.57 & 73.28 & 0.30 & 67.70 & \textbf{0.26} & 77.51 & 0.44 \\
& Dspy~\citep{DBLP:journals/corr/abs-2310-03714} & 80.90 & 0.55 & 80.90 & \textbf{0.60} & \textbf{76.86} & 0.34 & 63.60 & 0.21 & 76.99 & 0.44 \\
& \sys (CoST) & \textbf{88.03} & \textbf{0.72} & 80.45 & 0.56 & 76.72 & \textbf{0.37} & \textbf{68.35} & 0.25 & \textbf{79.01} & \textbf{0.48} \\
\midrule


\multirow{4}{*}{GPT-4o} 
& Zero-Shot & 84.10 & 0.73 & 80.53 & 0.60 & 81.50 & 0.50 & 64.30 & 0.25 & 79.32 & 0.54 \\
& CoT~\citep{DBLP:conf/nips/Wei0SBIXCLZ22} & 85.57 & 0.69 & 84.40 & \textbf{0.66} & 79.58 & 0.44 & 64.45 & \textbf{0.30} & 80.08 & 0.54 \\
& Dspy~\citep{DBLP:journals/corr/abs-2310-03714} & 86.17 & 0.74 & 82.65 & 0.62 & 80.00 & 0.47 & 67.25 & 0.27 & 80.26 & 0.54 \\
& \sys (CoST) & \textbf{88.80} & \textbf{0.75} & \textbf{84.50} & 0.65 & \textbf{82.24} & \textbf{0.50} & \textbf{68.90} & 0.28 & \textbf{82.39} & \textbf{0.56} \\


\bottomrule
\end{tabular}%
}
\vspace{-2em}
\end{table}

\subsection{Numerical Results of The Radar Chart}
\label{appendix:full_radar_results}

Table~\ref{tab:llm_enhancement} provides the detailed numerical results on the Finance subset of Loong across  backbone LLMs. Relative to Zero-Shot prompting, Chain-of-Thought (CoT) consistently improves overall performance, raising the average score (AS) from 75.60 to 77.51 on Qwen2.5-14B-Ins and from 79.32 to 80.08 on GPT-4o. Building further on CoT, our CoST-based \sys achieves the strongest results across all backbones, with overall AS/PR reaching 79.01/0.48 on Qwen2.5-14B-Ins, and 82.39/0.56 on GPT-4o. The improvements are consistent across subtasks—for example, CoST yields a gain of +6.62 AS on Clustering with GPT-4o and +8.62 AS on Spotlight Locating with Qwen2.5-14B-Ins, demonstrating its robustness in enhancing reasoning-intensive extraction.

We have also integrate prompt-optimization methods, including DSPy~\citep{DBLP:journals/corr/abs-2310-03714}, into our evaluation. Specifically, we use the same zero-shot structured-extraction instructions as the initial input prompts for GPT-4o and Qwen2.5-14B-Instruct, and then apply the MiPRO optimizer to refine these prompts—automatically generating Chain-of-Thought–style reasoning steps and optimized structured prompts. MiPRO is conditioned on examples drawn from a randomly sampled 10\% subset of the \sys training data (finance). We observe that although DSPy provides slight improvements over the zero-shot and CoT baselines, our method still performs substantially better. This highlights both the effectiveness and robustness of CoST, and further illustrates the challenges of applying automated prompt optimization to structure-aware extraction tasks.


\subsection{Robust Long-Document Handling.} 
\label{appendix:document-set}

\begin{figure}
    \centering
    \includegraphics[width=\linewidth]{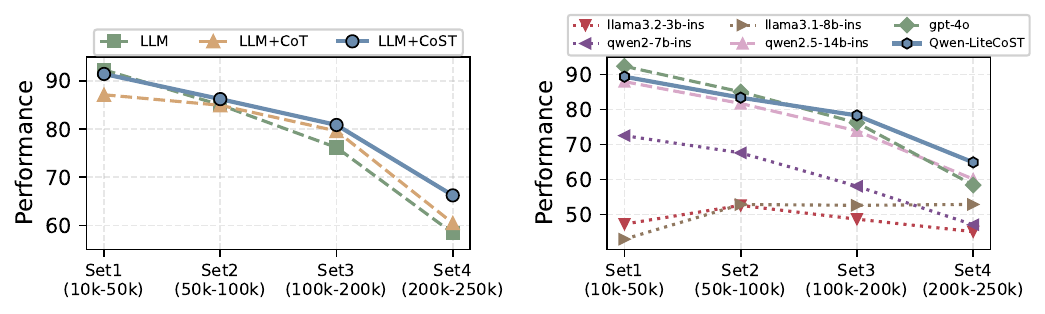}
    \vspace{-2em}
    \caption{Performance (LLM score) versus document-set size for (a) CoST with baseline methods (zero-shot and CoT) and (b) LiteCoST-tuned models with different base models.}
    \label{fig:document-set}
    \vspace{-1em}
\end{figure}

The results in Fig.~\ref{fig:document-set} show that both Chain-of-Structured-Thought (CoST) and \sys exhibit strong robustness as document length increases.

\textbf{LLM+CoST.} The overall score of LLM equipped with CoST decreases by only 25.21 points when moving from Set1 (91.46) to the most challenging Set4 (66.25), whereas zero-shot prompting exhibits a much sharper decline (92.33 $\rightarrow$ 58.42) and CoT similarly drops from 87.90 to 60.49. On Set4, CoST surpasses zero-shot and CoT by 5.76 and 7.83 points, respectively, indicating that CoST provides substantially greater robustness under increasing difficulty.

\textbf{\sys-tuned SLM.} Our \sys-tuned SLM also demonstrates strong robustness, exhibiting only a modest 24.46-point decline as document length increases. Even on the most challenging Set4, it achieves a score of 64.89, outperforming all LLM-based baselines: +6.47 over GPT-4o, +4.77 over Qwen2.5-14B-Instruct, and +12.01 over LLaMA-3.1-8B-Instruct.


These results underscore the particular strength of CoST and \sys in handling long-document inputs, consistently maintaining high reasoning accuracy even as context length increases.

\subsection{Full Results of Reasoning w/ Structured Data.}
\label{appendix: Loong_structure}

By selecting the optimal structure for each QA sample in the Loong benchmark, the dataset comprises 823 tables, 400 graphs, and 377 text chunks.
The purpose of this experiment is to explore how \textit{serialized structured output (SSO)} contributes to model performance on knowledge-intensive reasoning tasks. Therefore, we curate such structures through our \sys framework with GPT-4o as the base model, and then evaluate the results of LLMs reasoning over these structured representation.

Table~\ref{tab:ablation_sd} shows that all models achieve better reasoning performance when leveraging the extracted structured
information, compared to the baseline that directly processes the raw long-form documents, underscoring the necessity of accurate and effective structured knowledge extraction.

\textbf{Pros.} The gains are particularly notable in comparison and clustering tasks, where all models show improvements in LLM scores, along with at least 0.11 and 0.17 increases in Perfect Rate, respectively.
These results highlight the value of structured knowledge in capturing entity relationships and aggregating discrete information for complex reasoning.

\textbf{Cons.} We observe slight performance drops in certain subtasks (\textit{Spotlight Locating} and \textit{Chain-of-Reasoning}) after integrating structured data. These results reflect a trade-off between structured clarity and the flexibility of unstructured reasoning:
\begin{itemize}[topsep=1pt,itemsep=1pt,parsep=0pt]
    \item \textbf{Advantages of structured data:} Improves aggregation and clustering tasks by enabling more effective comparison and summarization.
    \item \textbf{Challenge in Spotlight Locating:} Fine-grained localization may degrade when structured data omits subtle contextual cues, reducing recall.
    \item \textbf{Challenge in Chain-of-Reasoning:} Complex multi-step inference may lose nuance when context is compressed into structured formats.
\end{itemize}

Here, we provide a case to illustrate this issue with the following Spotlight Locating task:
\vspace{-0.5em}

\begin{itemize}[leftmargin=*,topsep=1pt,itemsep=1pt,parsep=0pt]
    \item \textbf{Question:} What is the name of the company with $\$30,179$ in Accounts Receivable?
    \item \textbf{Ground Truth:} CIRTRAN CORP
    \item \textbf{LLM Answer:} The information provided in the context does not list a company with $\$30,179$ in Accounts Receivable. Therefore, the answer is $\boxed{\text{Not Provided}}$
\end{itemize}

\begin{table}[h]
\centering
\caption{Extracted table for the given query.}
\vspace{-1em}
\begin{tabular}{l c}
\toprule
\textbf{Company} & \textbf{Accounts Receivable} \\
\midrule
AXIM Biotechnologies, Inc. & \$23,642 \\
CIRTRAN CORP. & None \\
Arax Holdings Corp. & \$453,837 \\
High Wire Networks, Inc. & \$4,483,1 \\
\bottomrule
\end{tabular}
\end{table}

In this case, due to value corruption or omission during extraction, CIRTRAN CORP’s relevant numerical value is missing in the structured table, leading the model to respond incorrectly. A large language model performing direct reasoning over the full text, in contrast, can flexibly match partial evidence (e.g., approximate figures or nearby mentions) without being constrained by structure.

\begin{table}[t!]
\vspace{-1em}
\centering
\caption{Structured data quality evaluation of \sys via Chain-of-Structured-Thought (CoST) on the Loong benchmark, compared against popular LLMs. \textit{SD} denotes structured data.}
\vspace{-1em}

\label{tab:ablation_sd}
\setlength{\tabcolsep}{3pt} 
\small 
\resizebox{\textwidth}{!}{ 
\begin{tabular}{@{} l c *{10}{S[table-format=2.2]} @{}}
\toprule
\multirow{2}{*}{\makebox[1.8cm][l]{\textbf{Model}}} & \textbf{Context} & \multicolumn{2}{c}{\textbf{\textit{Spotlight Locating}}} & \multicolumn{2}{c}{\textbf{\textit{Comparison}}} & \multicolumn{2}{c}{\textbf{\textit{Clustering}}} & \multicolumn{2}{c}{\textbf{\textit{Chain of Reasoning}}} & \multicolumn{2}{c}{\textbf{\textit{Overall}}} \\
\cmidrule(lr){3-4} \cmidrule(lr){5-6} \cmidrule(lr){7-8} \cmidrule(lr){9-10} \cmidrule(lr){11-12}
 & \textbf{Length} & \textbf{\textit{AS}} & \textbf{\textit{PR}} & \textbf{\textit{AS}} & \textbf{\textit{PR}} & \textbf{\textit{AS}} & \textbf{\textit{PR}} & \textbf{\textit{AS}} & \textbf{\textit{PR}} & \textbf{\textit{AS}} & \textbf{\textit{PR}} \\
\midrule
Qwen2-72B-Ins & 128k & 54.17 & 0.36 & 42.38 & 0.20 & 36.71 & 0.04 & 47.76 & 0.18 & 43.29 & 0.15 \\
SD $\to$ Qwen2-72B-Ins & 128k & 57.30 & 0.31 & 54.80 & 0.38 & 61.46 & 0.23 & 46.36 & 0.16 & \colorbox{mygreen}{\textbf{55.70}} & \colorbox{mygreen}{\textbf{0.25}} \\
\midrule
GPT-4o-mini & 128k & 59.46 & 0.49 & 51.90 & 0.27 & 34.55 & 0.04 & 64.28 & 0.39 & 49.25 & 0.24 \\
SD $\to$ GPT-4o-mini & 128k & 63.23 & 0.44 & 53.04 & 0.38 & 59.63 & 0.21 & 55.98 & 0.26 & \colorbox{mygreen}{\textbf{58.02}} & \colorbox{mygreen}{\textbf{0.29}} \\
\midrule
GPT-4o & 128k & 73.95 & 0.62 & 50.50 & 0.28 & 44.29 & 0.09 & 57.05 & 0.28 & 53.47 & 0.26 \\
SD $\to$ GPT-4o & 128k & 62.11 & 0.33 & 63.27 & 0.41 & 68.06 & 0.29 & 53.52 & 0.22 & \colorbox{mygreen}{\textbf{62.51}} & \colorbox{mygreen}{\textbf{0.30}} \\
\midrule
Claude3.5-Sonnet & 200k & 58.45 & 0.49 & 54.21 & 0.25 & 45.77 & 0.07 & 43.92 & 0.25 & 48.85 & 0.23 \\
SD $\to$ Claude3.5-Sonnet & 200k & 47.60 & 0.34 & 54.64 & 0.41 & 66.95 & 0.31 & 50.45 & 0.23 & \colorbox{mygreen}{\textbf{57.32}} & \colorbox{mygreen}{\textbf{0.31}} \\
\bottomrule
\end{tabular}%
}
\vspace{-1em}
\end{table}


\subsection{Computational Resource.}
\label{appendix:computational_resource}
\vspace{-0.5em}

\textbf{Model Deployment.} We deployed and run models ranging from 3B to 14B model size on a cluster equipped with eight NVIDIA RTX 4090 GPUs, each with 24GB of VRAM. For closed sourced large language models such as GPT and Claude, we accessed them via API calls.

\textbf{Computational cost.} Our cost mainly comes from two components: 
\vspace{-0.5em}

\begin{itemize}[leftmargin=*, topsep=1pt,itemsep=1pt,parsep=0pt]
    \item \textbf{Data Generation.} This includes structure analysis, CoST trace generation, quality verification, and iterative refinement using GPT-4o. The average cost is approximately \textbf{\$30 }per domain.
    \item \textbf{GRPO Fine-tuning.} Reward computation relies on GPT-4o-mini to evaluate structural alignment, format compliance, and answer correctness. This adds an additional \textbf{~\$10} per training run.
\end{itemize}

The total cost of \textbf{~\$40} is necessary and acceptable because it amortizes extremely well: once LiteCoST is trained, downstream inference relies solely on compact SLMs, eliminating repeated LLM calls and yielding substantial savings during deployment.


\section{Case Study on Practical Long-Document QA Tasks.}
\label{appendix:case_study}
\vspace{-0.5em}

\subsection{Case Study on RL-Enhancement}
\label{appendix:case_study_rlenhancement}
\begin{figure}
    \centering
    \includegraphics[width=0.95\linewidth]{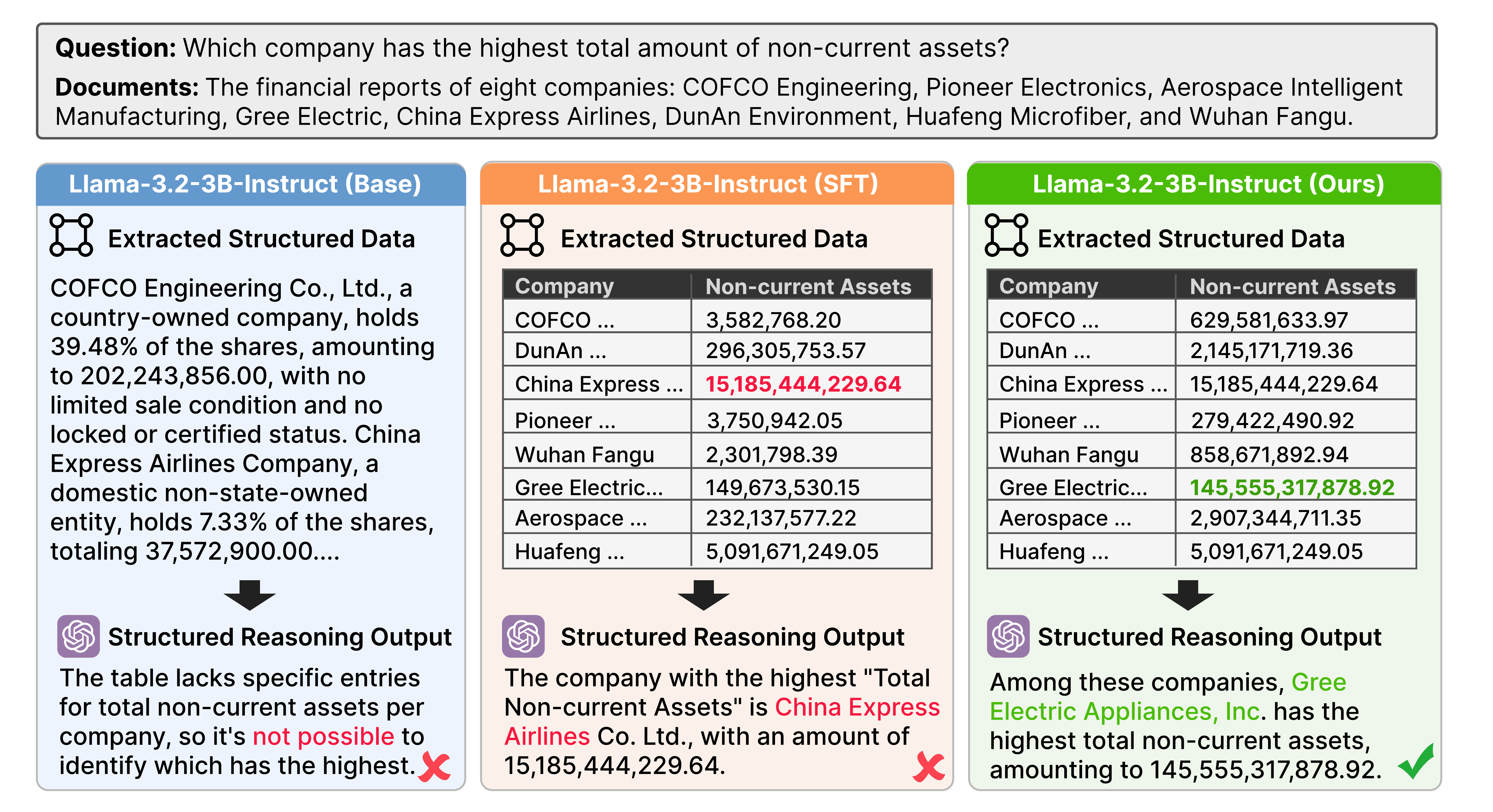}
    \vspace{-1em}
    \caption{Case study on representative information extraction tasks for large language models, comparing the information extraction capabilities of three Llama-3.2-3B-Instruct variants.}
    \label{fig:case_study}
    \vspace{-1em}
\end{figure}

As shown in Fig.~\ref{fig:case_study}, we present a representative long-context QA example to evaluate the model's information extraction capabilities. This question is particularly challenging, as it requires the model to retrieve multiple pieces of evidence from the input text and accurately integrate them into coherent structured information.
Specifically, the case study includes the question and documents input (grey box), predictions from the base LLaMA-3.2-3B-Instruct model (blue box), its finetuned variant (orange box) and our \sys model (green box). 

In addition, incorrect predictions are highlighted in red, while correct ones are marked in green. The results show that the Llama base model performs poorly at information extraction and fails to organize content into structured formats such as tables. The fine-tuned model exhibits partial extraction ability but remains inaccurate in long-context, multi-document settings. In contrast, our \sys accurately integrates dispersed information into high-quality structured tables, with gains largely attributed to supervised CoT reasoning and reinforcement learning. As shown in Fig.~\ref{fig:case_study}, \sys provides significantly enhanced interpretability compared to its base model.

\section{Generalization on Different Domains}
\label{appendix:generalization}

\subsection{Legal Domain.}
\label{appendix:legal_qa}

To examine the generalization capabilities of \sys, we apply it to the legal domain by curated by from LegalBenchRAG~\citep{pipitone2024legalbench}, a dataset
of 6,858 query-answer pairs over a corpus of over 79M characters, entirely human-annotated by legal experts.
Leveraging the training data described in Sec.~\ref{ssec:exp_setup}, we perform RL fine-tuning on small language models. The fine-tuned models are subsequently evaluated on the legal subset of Loong~\citep{DBLP:conf/emnlp/WangCCL0WYXZLLY24} to assess their ability to generate structured outputs in complex legal contexts.



Table~\ref{tab:loonglegal} shows that our \sys-tuned 3B/7B models, LLaMA-LiteCoST and Qwen-LiteCoST, achieve substantial improvements over their respective base models and deliver performance comparable to proprietary LLMs. From these results, we further derive two key insights:

(1) Notably, our method elevates LLaMA variant to surpass Qwen in overall performance. The two backbones, however, exhibit complementary strengths: LLaMA excels on the Spotlight Locating and Comparison subtasks, while Qwen achieves superior results on Clustering and Chain-of-Reasoning.  

(2) Qwen2-7B-Instruct performs exceptionally well on the legal domain, even surpassing its 14B counterpart, potentially due to:
\noindent\ding{182} stronger domain adaptation from a larger proportion of legal corpora in training;
\noindent\ding{183} the tendency of larger models to over-generate in long, highly structured legal texts, leading to hallucinations or format drift, whereas the smaller 7B model more faithfully adheres to schemas and maintains consistency.

\subsection{Open-domain QA.}
\label{appendix:scientific_qa}

Beyond the Loong benchmark, we further evaluate our method on LongBench~\citep{bai2024longbench}, a widely recognized multi-task benchmark for long-document QA that covers key real-world application scenarios across literature, science, encyclopedias, etc. These datasets contain far more nuanced information. As shown in Table~\ref{tab:combined_qa}, our \sys effectively extends to settings with richer and more subtle semantics, demonstrating strong generalization beyond strictly structured environments. Our analysis focuses on both single- and multi-doc QA tasks across four datasets, as shown in Table~\ref{tab:longbench_overview}.

\textbf{Single-Doc QA.} For single-document QA, we focus on datasets containing longer and more challenging documents. We evaluate on NarrativeQA~\citep{kovcisky2018narrativeqa}, consisting of full-length stories paired with questions designed to test deep reading comprehension. We also include Qasper~\citep{dasigi2021dataset}, a dataset featuring QA over NLP research papers, annotated by domain experts.

\textbf{Multi-Doc QA.} Multi-document QA requires models to extract and combine information from several
documents to obtain the answer, which is usually
more challenging than single-doc QA. We evaluate on two  Wikipedia-based multi-hop QA datasets: HotpotQA~\citep{yang2018hotpotqa}, 2WikiMultihopQA~\citep{DBLP:conf/coling/HoNSA20}. HotpotQA involves a number of 2-hop questions directly written by na-
tive speakers given two related paragraphs. 
2Wiki-MultihopQA consists of up to 5-hop questions that
are synthesized through manually designed tem-
plates to ensure that they cannot be solved through
shortcuts. Each question in the original datasets
is supplemented by 2-4 supporting paragraphs that
provide one-step reasoning evidence and several
distracting paragraphs.

\begin{table*}[t]
\centering  
\small
\caption{An overview of the dataset statistics for a subset of LongBench. `Source' denotes the origin of the context. `Avg len' (average length) is computed using the number of words for the English (code) datasets and the number of characters for the Chinese datasets. `Accuracy (CLS)' refers to classification accuracy, while `Accuracy (EM)' refers to exact match accuracy.}
\label{tab:longbench_overview}
\resizebox{\textwidth}{!}{
\begin{tabular}{lclrccc}
\toprule
Dataset & ID & Source & Avg len & Metric & Language & \#data \\
\midrule
\emph{Single-Document QA} \\
NarrativeQA & 1-1 & Literature, Film & 18,409 & F1 & English & 200 \\
Qasper & 1-2 & Science & 3,619 & F1 & English & 200 \\
\midrule
\emph{Multi-Document QA} \\
HotpotQA & 2-1 & Wikipedia & 9,151 & F1 & English & 200 \\
2WikiMultihopQA & 2-2 & Wikipedia & 4,887 & F1 & English & 200 \\
\bottomrule
\end{tabular}
}
\label{tb:stat}
\vspace{-5mm}
\end{table*}

\subsection{Others.}
\label{appendix:other_domain}

Beyond the finance, legal, and scientific QA tasks, \sys can be applied to a wide range of fields, enabling effective and efficient structured extraction from large-scale unstructured corpora:

\begin{itemize}[noitemsep, topsep=0pt, leftmargin=1.5em]
    \item \textbf{Healthcare}: extracting patient attributes, treatment outcomes, and adverse event reports to enhance clinical decision-making and pharmacovigilance.  
    \item \textbf{Scientific literature}: supporting literature understanding, analysis, and question answering through the extraction of experimental settings, results, and methodological details.  
    \item \textbf{Policy and government}: structuring entities and relations from legislative and regulatory documents to facilitate compliance monitoring and policy evaluation.  
    \item \textbf{Enterprise analytics}: organizing information from reports, manuals, and support logs into structured forms to improve knowledge management and retrieval-augmented applications.  
\end{itemize}

Together, these applications highlight the broad adaptability of \sys in transforming unstructured knowledge into structured representations that directly support diverse downstream tasks.

\section{Full Performance of Effectiveness on Loongfin}
\label{appendix:full_result}
Table~\ref{tab:full_results} presents the full results of the Table~\ref{tab:litesea_results} in the main manuscript.


\begin{table*}[t]
\centering  
\caption{Full results of Table~\ref{tab:litesea_results}. \textit{AS} represents \textit{Avg Scores (0\textasciitilde100)} and \textit{{PR}} denotes \textit{Perfect Rate} (0\textasciitilde1). \textbf{Bold} indicates the best result within each setting, and \underline{underlined} indicates the second best.}
\label{tab:full_results}
\renewcommand{\arraystretch}{1}
\resizebox{\textwidth}{!}{
\begin{tabular}{llcccccccccc}
\toprule
\multirow{2}{*}{\textbf{Model}} & \multicolumn{1}{c}{\textbf{Model}} & \multicolumn{2}{c}{\textbf{Spotlight Locating}} & \multicolumn{2}{c}{\textbf{Comparison}} & \multicolumn{2}{c}{\textbf{Clustering}} & \multicolumn{2}{c}{\textbf{Chain of Reasoning}} & \multicolumn{2}{c}{\textbf{Overall}}\\ \cmidrule(r){3-4} \cmidrule(r){5-6} \cmidrule(r){7-8} \cmidrule(r){9-10} \cmidrule(r){11-12}
 & \multicolumn{1}{c}{\textbf{Size}} & \textbf{\textit{AS}} & \textbf{\textit{PR}} & \textbf{\textit{AS}} & \textbf{\textit{PR}} & \textbf{\textit{AS}} & \textbf{\textit{PR}} & \textbf{\textit{AS}} & \textbf{\textit{PR}} & \textbf{\textit{AS}} & \textbf{\textit{PR}} \\

\midrule
\multicolumn{12}{>{\columncolor[gray]{.88}}c}{\textbf{$\mathtt{All\ Set}$ (10K-250K)}}  \\
LLaMA-3.2-3B-Instruct & 3B & 49.90 & 0.16 & 52.10 & 0.14 & 47.89 & 0.07 & 46.85 & 0.06 & 49.37 & 0.11 \\ 
Qwen2-7B-Instruct & 7B & 63.10 & 0.36 & 67.85 & 0.37 & 60.83 & 0.18 & 52.25 & 0.09 & 62.10 & 0.26 \\
LLaMA-3.1-8B-Instruct & 8B & 55.03 & 0.20 & 51.60 & 0.15 & 51.50 & 0.04 & 44.75 & 0.02 & 51.32 & 0.10 \\
GPT-4o-mini & 8B & 84.42 & 0.70 & 80.40 & 0.67 & 77.38 & 0.40 & 65.35 & 0.18 & 78.08 & \underline{0.51} \\
Qwen2.5-14B-Instruct & 14B & 83.74 & 0.57 & 82.12 & 0.56 & 69.96 & 0.24 & 66.41 & 0.10 & 75.60 & 0.38 \\
GPT-4o & 200B & 84.10 & 0.73 & 80.53 & 0.60 & 81.50 & 0.50 & 64.30 & 0.25 & \underline{79.32} & \textbf{0.54} \\
Deepseek-R1  & 671B & 84.27 & 0.62 & 78.97 & 0.55 & 75.42 & 0.34 & 74.40 & 0.35 & 78.18 & 0.46 \\
\midrule
LLaMA-3.2-3B-Instruct (\textit{SFT}) & 3B & 74.39 & 0.45 & 75.53 & 0.45 & 73.64 & 0.29 & 59.15 & 0.12 & 72.27 & 0.35 \\ 
LLaMA-3.2-3B-Instruct (\textit{Ours}) & 3B & 81.27 & 0.53 & 78.08 & 0.49 & 78.34 & 0.36 & 64.75 & 0.16 & 76.95 & 0.40 \\
Qwen2-7B-Instruct (\textit{SFT}) & 7B & 82.23 & 0.58 & 81.15 & 0.56 & 75.91 & 0.33 & 62.40 & 0.11 & 76.83 & 0.42 \\ 
Qwen2-7B-Instruct (\textit{Ours}) & 7B & 83.97 & 0.62 & 81.55 & 0.59 & 81.00 & 0.43 & 67.98 & 0.18 & \textbf{79.93} & 0.48 \\

\midrule
\multicolumn{12}{>{\columncolor[gray]{.88}}c}{\textbf{$\mathtt{Set1}$ (10K-50K)}}  \\
LLaMA-3.2-3B-Instruct & 3B & 54.13 & 0.17 & 43.33 & 0.13 & 44.25 & 0.07 & 55.50 & 0.10 & 47.28 & 0.12 \\ 
Qwen2-7B-Instruct & 7B & 73.26 & 0.48 & 80.17 & 0.53 & 68.25 & 0.28 & 65.00 & 0.30 & 72.52 & 0.40 \\
LLaMA-3.1-8B-Instruct & 8B & 46.09 & 0.04 & 35.50 & 0.00 & 47.38 & 0.00 & 40.50 & 0.00 & 42.96 & 0.01 \\ 
GPT-4o-mini & 8B & 96.09 & 0.91 & 93.00 & 0.90 & 86.62 & 0.70 & 75.50 & 0.40 & 89.51 & \underline{0.78} \\ 
Qwen2.5-14B-Instruct & 14B & 95.00 & 0.74 & 87.00 & 0.63 & 84.45 & 0.53 & 84.30 & 0.20 & 87.53 & 0.57 \\ 
GPT-4o & 200B & 96.09 & 0.91 & 90.00 & 0.80 & 91.25 & 0.78 & 95.00 & 0.90 & \textbf{92.33} & \textbf{0.83} \\ 
Deepseek-R1  & 671B & 91.96 & 0.78 & 90.50 & 0.80 & 88.75 & 0.68 & 99.50 & 0.90 & \underline{91.02} & 0.76 \\ 
\midrule
LLaMA-3.2-3B-Instruct (\textit{SFT}) & 3B & 80.43 & 0.61 & 83.10 & 0.60 & 82.70 & 0.47 & 67.00 & 0.30 & 80.79 & 0.52 \\
LLaMA-3.2-3B-Instruct (\textit{Ours}) & 3B & 91.52 & 0.78 & 81.33 & 0.57 & 88.95 & 0.60 & 75.00 & 0.50 & 85.95 & 0.62 \\ 
Qwen2-7B-Instruct (\textit{SFT}) & 7B & 91.30 & 0.83 & 89.17 & 0.73 & 81.75 & 0.53 & 81.50 & 0.20 & 86.02 & 0.62 \\ 
Qwen2-7B-Instruct (\textit{Ours}) & 7B & 92.17 & 0.83 & 89.67 & 0.77 & 88.75 & 0.72 & 84.80 & 0.60 & 89.40 & 0.75 \\

\midrule
\multicolumn{12}{>{\columncolor[gray]{.88}}c}{\textbf{$\mathtt{Set2}$ (50K-100K)}}  \\
LLaMA-3.2-3B-Instruct & 3B & 41.62 & 0.10 & 57.16 & 0.19 & 57.13 & 0.13 & 44.88 & 0.05 & 52.61 & 0.13 \\ 
Qwen2-7B-Instruct & 7B & 80.00 & 0.62 & 71.13 & 0.45 & 63.11 & 0.24 & 58.75 & 0.12 & 67.61 & 0.35 \\
LLaMA-3.1-8B-Instruct & 8B & 55.88 & 0.23 & 55.15 & 0.20 & 51.44 & 0.03 & 48.75 & 0.03 &  52.86 & 0.11 \\ 
GPT-4o-mini & 8B & 96.12 & 0.90 & 88.93 & 0.80 & 86.07 & 0.58 & 64.62 & 0.28 & \textbf{85.09} & \underline{0.65} \\ 
Qwen2.5-14B-Instruct & 14B & 88.88 & 0.68 & 88.67 & 0.68 & 76.11 & 0.28 & 64.25 & 0.20 & 80.10 & 0.45 \\ 
GPT-4o & 200B & 90.38 & 0.80 & 96.27 & 0.72 & 89.67 & 0.71 & 66.88 & 0.33 & \underline{85.02} & \textbf{0.67} \\ 
Deepseek-R1  & 671B & 85.75 & 0.68 & 83.04 & 0.61 & 81.17 & 0.50 & 78.62 & 0.45 & 82.07 & 0.56 \\
\midrule
LLaMA-3.2-3B-Instruct (\textit{SFT}) & 3B & 82.83 & 0.57 & 77.93 & 0.49 & 79.09 & 0.40 & 65.12 & 0.15 & 77.07 & 0.42 \\
LLaMA-3.2-3B-Instruct (\textit{Ours}) & 3B & 88.12 & 0.60 & 81.33 & 0.56 & 83.98 & 0.48 & 65.25 & 0.20 & 80.79 & 0.40 \\ 
Qwen2-7B-Instruct (\textit{SFT}) & 7B & 87.88 & 0.70 & 84.73 & 0.63 & 80.53 & 0.41 & 66.12 & 0.20 & 80.67 & 0.40 \\ 
Qwen2-7B-Instruct (\textit{Ours}) & 7B & 90.00 & 0.72 & 85.53 & 0.68 & 85.28 & 0.56 & 68.50 & 0.23 & 83.39 & 0.50 \\ 

\midrule
\multicolumn{12}{>{\columncolor[gray]{.88}}c}{\textbf{$\mathtt{Set3}$ (100K-200K)}}  \\
LLaMA-3.2-3B-Instruct & 3B & 54.42 & 0.22 & 51.24 & 0.09 & 44.06 & 0.02 & 45.14 & 0.09 & 48.67 & 0.10 \\ 
Qwen2-7B-Instruct & 7B & 56.00 & 0.30 & 63.73 & 0.28 & 59.63 & 0.13 & 45.57 & 0.03 & 58.08 & 0.20 \\
LLaMA-3.1-8B-Instruct & 8B & 57.92 & 0.27 & 55.13 & 0.17 & 52.06 & 0.04 & 39.71 & 0.00 & 52.63 & 0.13 \\ 
GPT-4o-mini & 8B & 84.00 & 0.77 & 77.40 & 0.57 & 71.78 & 0.22 & 64.57 & 0.09 & 75.25 & \underline{0.43} \\ 
Qwen2.5-14B-Instruct & 14B & 88.93 & 0.68 & 76.93 & 0.49 & 63.39 & 0.16 & 64.14 & 0.00 & 73.29 & 0.35 \\ 
GPT-4o & 200B & 88.33 & 0.83 & 76.13 & 0.49 & 77.00 & 0.32 & 53.43 & 0.06 & 76.19 & \textbf{0.45} \\ 
Deepseek-R1  & 671B & 91.17 & 0.73 & 76.80 & 0.47 & 71.22 & 0.16 & 67.57 & 0.20 & \underline{76.94} & 0.38 \\
\midrule
LLaMA-3.2-3B-Instruct (\textit{SFT}) & 3B & 75.25 & 0.48 & 74.24 & 0.40 & 67.67 & 0.16 & 55.14 & 0.09 & 69.63 & 0.29 \\ 
LLaMA-3.2-3B-Instruct (\textit{Ours}) & 3B & 83.75 & 0.60 & 77.73 & 0.47 & 72.37 & 0.22 & 62.43 & 0.06 & 75.20 & 0.36 \\ 
Qwen2-7B-Instruct (\textit{SFT}) & 7B & 85.33 & 0.63 & 78.60 & 0.48 & 74.06 & 0.24 & 56.29 & 0.03 & 75.58 & 0.37 \\ 
Qwen2-7B-Instruct (\textit{Ours}) & 7B & 86.92 & 0.68 & 8.73 & 0.53 & 79.11 & 0.33 & 63.86 & 0.11 & \textbf{78.75} & 0.40 \\ 

\midrule
\multicolumn{12}{>{\columncolor[gray]{.88}}c}{\textbf{$\mathtt{Set4}$ (200K-250K)}}  \\
LLaMA-3.2-3B-Instruct & 3B & 48.52 & 0.11 & 49.50 & 0.15 & 36.50 & 0.00 & 50.33 & 0.00 & 45.11 & 0.07 \\ 
Qwen2-7B-Instruct & 7B & 45.19 & 0.00 & 52.50 & 0.15 & 47.67 & 0.00 & 42.00 & 0.00 & 47.07 & 0.03 \\
LLaMA-3.1-8B-Instruct & 8B & 55.00 & 0.15 & 49.25 & 0.15 & 55.50 & 0.07 & 48.67 & 0.07 & 52.88 & 0.11 \\ 
GPT-4o-mini & 8B & 58.07 & 0.07 & 41.50 & 0.20 & 55.83 & 0.03 & 63.67 & 0.00 & 54.65 & 0.08 \\ 
Qwen2.5-14B-Instruct & 14B & 55.55 & 0.00 & 59.75 & 0.25 & 51.93 & 0.00 & 65.53 & 0.00 & 56.75 & 0.05 \\ 
GPT-4o & 200B & 55.19 & 0.22 & 61.25 & 0.25 & 57.50 & 0.03 & 62.33 & 0.07 & 58.42 & \textbf{0.14} \\ 
Deepseek-R1  & 671B & 60.19 & 0.15 & 54.50 & 0.25 & 53.00 & 0.00 & 62.33 & 0.07 & 56.96 & 0.11 \\
\midrule
LLaMA-3.2-3B-Instruct (\textit{SFT}) & 3B & 54.81 & 0.07 & 60.00 & 0.20 & 63.17 & 0.13 & 47.33 & 0.00 & 57.45 & 0.11 \\ 
LLaMA-3.2-3B-Instruct (\textit{Ours}) & 3B & 56.89 & 0.04 & 2.25 & 0.20 & 65.17 & 0.10 & 62.00 & 0.07 & \underline{61.59} & 0.10 \\ 
Qwen2-7B-Instruct (\textit{SFT}) & 7B & 59.26 & 0.07 & 65.25 & 0.30 & 59.83 & 0.10 & 54.00 & 0.00 & 59.89 & 0.12 \\ 
Qwen2-7B-Instruct (\textit{Ours}) & 7B & 61.48 & 0.15 & 67.75 & 0.30 & 66.67 & 0.07 & 65.33 & 0.00 & \textbf{65.16} & \underline{0.13} \\

\bottomrule
\end{tabular}
}

\end{table*}

\end{document}